\documentclass[11pt]{article}
\usepackage[square,numbers,sort]{natbib}
\usepackage[left=1in,right=1in,top=1in,bottom=1in]{geometry}

\usepackage[utf8]{inputenc} 
\usepackage[T1]{fontenc}    
\usepackage[colorlinks=true,citecolor=blue,urlcolor=blue]{hyperref}       
\usepackage{url}            
\usepackage{booktabs}       
\usepackage{amsfonts}       
\usepackage{nicefrac}       
\usepackage{microtype}      
\usepackage[dvipsnames]{xcolor}         

\usepackage{xspace} 
\usepackage{enumitem} 
\usepackage{multirow}
\usepackage{graphicx}
\usepackage[labelfont=bf,font=normalsize]{caption}
\usepackage{subcaption}
\usepackage{wrapfig}
\usepackage{amsmath,amssymb,amsthm}
\usepackage{cleveref}
\usepackage{mathtools}
\usepackage{algorithm}
\usepackage[algo2e, ruled, linesnumbered]{algorithm2e}
\usepackage{algorithmic}
\usepackage{dsfont}
\usepackage[skip=5pt plus1pt, indent=20pt]{parskip}

\makeatletter
\newcommand\footnoteref[1]{\protected@xdef\@thefnmark{\ref{#1}}\@footnotemark}
\makeatother

\DeclarePairedDelimiterX{\infdivx}[2]{(}{)}{%
  #1\;\delimsize\|\;#2%
}

\title{Profit: Benchmarking Personalization and Robustness Trade-off in Federated Prompt Tuning}
\author{Liam Collins\thanks{The University of Texas at Austin, work done while an intern at Google},\;\; Shanshan Wu\thanks{Google},\;\; Sewoong Oh$^{\dagger,}$\thanks{The University of Washington},\;\; Khe Chai Sim$^\dagger$ \\
\texttt{liamc@utexas.edu\;\; \{shanshanw, sewoongo, khechai\}@google.com} \\
}

\date{}
\begin{document}

\maketitle

\begin{abstract}
In many applications of federated learning (FL), clients desire models that are personalized using their local data, yet are also robust in the sense that they retain general global knowledge. However, the presence of data heterogeneity across clients induces a fundamental trade-off between personalization (i.e., adaptation to a local distribution) and robustness (i.e., not forgetting previously learned general knowledge). It is critical to  understand how to navigate this personalization vs robustness trade-off when designing federated systems, which are increasingly moving towards a paradigm of fine-tuning large foundation models. Due to limited computational and communication capabilities in most federated settings, this foundation model fine-tuning must be done using parameter-efficient fine-tuning (PEFT) approaches. While some recent work has studied federated approaches to PEFT, the personalization vs robustness trade-off of federated PEFT has been largely unexplored. In this work, we take a step towards bridging this gap by benchmarking fundamental FL algorithms -- FedAvg and FedSGD plus personalization (via client local fine-tuning) -- applied to one of the most ubiquitous PEFT approaches to large language models (LLMs) -- prompt tuning -- in a multitude of hyperparameter settings under varying levels of data heterogeneity.
Our results show that federated-trained prompts can be surprisingly robust when using a small learning rate with many local epochs for personalization, especially when using an adaptive optimizer as the client optimizer during federated training. We also demonstrate that simple approaches such as adding regularization and interpolating two prompts are effective in improving the personalization vs robustness trade-off in computation-limited settings with few local updates allowed for personalization.
\end{abstract}

\section{Introduction}

Federated learning (FL) is a framework that enables distributed clients to collaboratively train machine learning models in a privacy-preserving manner~\citep{mcmahan2017communication, kairouz2021advances, lireviewpaper, yang2019federated}. Unlike traditional server-side distributed training, in FL, each client (e.g., a mobile device)'s local data may follow a distinct distribution. This data heterogeneity motivates the development of personalized FL: the goal is to learn client-specific models that work well for each client's own data. Among all the personalized FL approaches~\cite[e.g.,][]{tan2022towards, wang2021field, wu2022motley, chen2022pfl}, one of the simplest methods is fine-tuning a global model on each client's local data to produce a personalized model~\citep{yu2020salvaging,jiang2019improving}. Despite its simplicity, fine-tuning a FedAvg (Federated Averaging~\citep{reddi2021adaptive, mcmahan2017communication})-trained global model has connections to meta learning~\citep{jiang2019improving, charles2023towards} and representation learning~\citep{collins2022fedavg}, and has been shown to work well over on-device data~\citep{wang2019federated, sim2021robust}. 

\begin{wrapfigure}[16]{r}{0.45\textwidth}
     \centering
     \includegraphics[width=\linewidth]{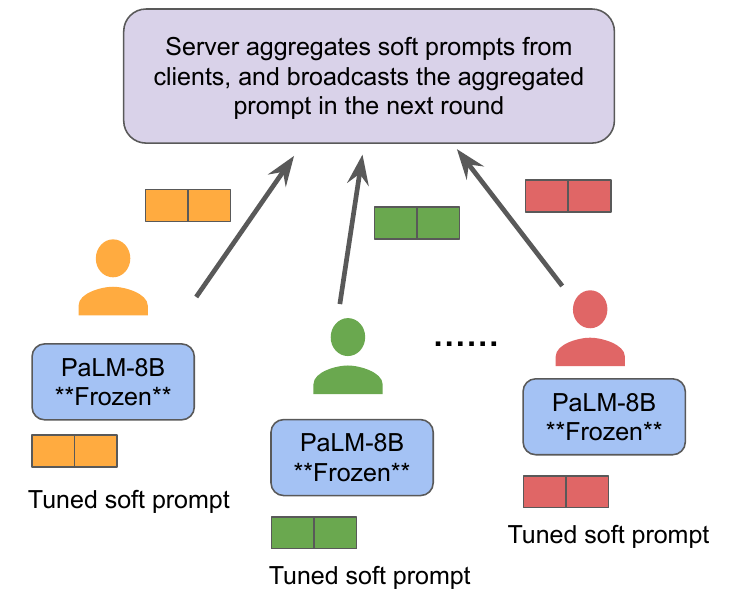}
     \caption{In each training round, only the soft prompts are updated and communicated between server and clients.}
     \label{fig:fed_prompt_tuning}
\end{wrapfigure}

Most of the existing FL personalization benchmarks (e.g.,~\cite{wu2022motley,chen2022pfl,matsuda2022empirical}) focus on training small-sized models (e.g., in the order of 10M parameters) from scratch. In this paper, we consider prompt tuning a pre-trained large language model (LLM) (specifically, an 8B parameter version of the PaLM model~\cite{chowdhery2022palm}) in the federated setting. As shown in Figure~\ref{fig:fed_prompt_tuning}, similar to the setup considered in~\cite{zhang2023fedyolo}, during FedAvg training, the PaLM-8B model is kept frozen, 
and only the soft prompt part is tuned and communicated between the server and clients; and during the personalization phase, each client will fine-tune the soft prompt locally to create a personalized soft prompt. Prompt tuning~\cite{lester2021power} is one of the standard parameter-efficient fine-tuning (PEFT) algorithms~\cite{ding2022delta,lialin2023scaling} proposed for LLMs. Considering the potential communication and memory limitations in the FL settings, PEFT is more suitable than full-model fine-tuning; besides, PEFT is shown to be capable of matching full-model fine-tuning in many scenarios~\cite{lester2021power,hu2021lora}. To create a federated dataset, similar to~\cite{zhang2023towards}, we partition a large-scale instruction tuning dataset based on the task types. We create datasets with three different heterogeneity levels (see Figure~\ref{fig:experimental_procedure} for an overview of our setup).

Our contributions are summarized below:

\begin{itemize}[leftmargin=*]
    \item We run comprehensive experiments to study the trade-off between personalization (adaptation to the clients' local distributions) and robustness (not forgetting the previously learned knowledge obtained during the FL training) over different FL training algorithms (variants of FedAvg and FedSGD) and different data heterogeneity levels (high/medium/low). To our knowledge, we are the first to study this trade-off in the setting of FL personalization and LLM prompt tuning. 
    \item We observe that for federated prompt tuning, it is important to use an adaptive optimizer (e.g., Adam~\cite{kingma2014adam}) as the client optimizer\footnote{\label{stateless}Note that the resulting algorithm is still a \emph{stateless} algorithm. A stateless algorithm means that the client does not maintain states locally and reuse them in the next participating round~\cite{kairouz2021advances,wang2021field, wu2022motley}. In our setting, it means that clients do not store Adam optimizer state (estimates of moments). Stateful algorithms (e.g., SCAFFOLD~\cite{karimireddy2020scaffold}) can perform poorly with low clients participating rate (see Section 5.1 of~\cite{reddi2021adaptive}).} in FedAvg (even though the server optimizer already uses adaptive optimizer). This is unlike the previously proposed adaptive FedAvg algorithm~\cite{reddi2021adaptive} (which uses adaptive optimizer at the server, and vanilla SGD at the clients). Our hypothesis is that the loss surface is very flat relative to the large scale of the learned soft prompt, so using an adaptive optimizer at the clients are crucial in making enough progress during training (see Section~\ref{sec:experiments} Observation 3a).
    \item We observe that during the personalization stage (i.e., during the local prompt fine-tuning stage), reducing the learning rate improves the personalization vs robustness trade-off, but it necessitates running many steps to reach the best personalization performance. We find that simple methods such as adding $\ell_2$ regularization and/or model averaging are effective to achieve the best of both worlds: better personalization vs robustness trade-off in fewer local tuning steps (see Figure \ref{fig:heuristics}).
\end{itemize}

\section{Related Works}

\textbf{Federated PEFT of pre-trained LLMs.} 
A number of works have begun to explore PEFT in the federated settings. Some have studied federated prompt tuning on vision tasks, without evaluating personalization \citep{zhang2023gpt,chen2022fedtune,guo2023promptfl}. Other works have benchmarked federated PEFT on language tasks, but again did not consider personalization \citep{zhang2023towards,zhao2023fedprompt,cai2022aug,cai2022fedadapter}. To our knowledge, all studies of federated PEFT that  consider personalization focus on the vision modality \citep{guo2023pfedprompt,li2023visual,lu2023fedclip,su2022cross,zhang2023fedyolo}.
Outside of PEFT, 
 \cite{hilmkil2021scaling,tian2022fedbert,weller2022pretrained} studied federated full-model fine-tuning of BERT models, which are at least an order of magnitude smaller than modern LLMs. Multiple works have noticed that initializing full-model federated training from a pre-trained model can mitigate the effects of data heterogeneity \citep{nguyen2022begin,weller2022pretrained,chen2022pre}.
Like our work, \cite{nguyen2022begin} also noticed the importance of using adaptive optimizers when running federated fine-tuning, but they only considered full-model fine-tuning starting from small models. 
Other works have analyzed the effect of differential privacy on federated training of language models via initialization with \cite{li2021large} or by distillation from a pre-trained LLM \cite{wang2023can} .

\textbf{Personalization in FL.} 
A long line of work within federated learning has developed techniques for personalizing models to each client
\citep{deng2020adaptive,hanzely2020federated,t2020personalized,fallah2020personalized,li2021ditto,mansour2020three,singhal2021federated,collins2021exploiting,shamsian2021personalized,marfoq2022personalized}.
We defer readers to the recent FL personalization benchmarks~\cite{wu2022motley,chen2022pfl,matsuda2022empirical} and the references therein for a more detailed discussion of the related work. In this paper, we focus on one of the simplest personalization approaches: each client fine-tunes a model locally to get the personalized model \citep{yu2020salvaging,jiang2019improving,collins2022fedavg,cheng2021fine,charles2023towards}. In particular, we are interested in studying the personalization and robustness trade-off. To our knowledge, we are the first to study this trade-off in the setting of federated prompt tuning for LLMs.








\textbf{Robustness to catastrophic forgetting during fine-tuning.} Robustness can have different definitions, e.g., robustness to attacks~\cite{li2021ditto, wang2021robust} and outliers~\cite{kundu2022robust}. In this paper, we focus on a special type, that is, robustness to forgetting about the global knowledge learned by FedAvg when each client fine-tunes the global prompt locally to get a personalized prompt. This is connected to the robustness to distribution shift or out-of-distribution data in the literature, see, e.g.,~\cite{andreassen2021evolution, wortsman2022model, wortsman2022robust, ilharco2022patching, tran2022plex, kumar2022fine, jiang2023test}, where the main difference is that in our experiments, the in-distribution and out-of-distribution have a special connection unique to the FL setting: a client's local distribution vs all clients' joint distribution.
Catastrophic forgetting~\cite{mccloskey1989catastrophic} has been studied for decades. Many proposed methods (e.g.,~\cite{rolnick2019experience, kirkpatrick2017overcoming}) may not directly fit the FL setting due to privacy or computation constraint. \cite{sim2021robust} considers a production FL scenario, and proposes to let each client to decide whether to accept the personalized model based on validation data metric. This is orthogonal to the robust fine-tuning methods we experiment with in Figure~\ref{fig:heuristics}, where we tried two simple robust fine-tuning methods (regularization and model averaging~\cite{wortsman2022model, wortsman2022robust, ilharco2022patching}) that do not modify model architecture. We leave the investigation of more complicated robust fine-tuning methods (e.g., ~\cite{tran2022plex, jiang2023test}) to future work. 

\section{Experimental Setup} \label{sec:experiment_setup}
\begin{figure}[t]
    \centering
    \includegraphics[width=0.9\linewidth]{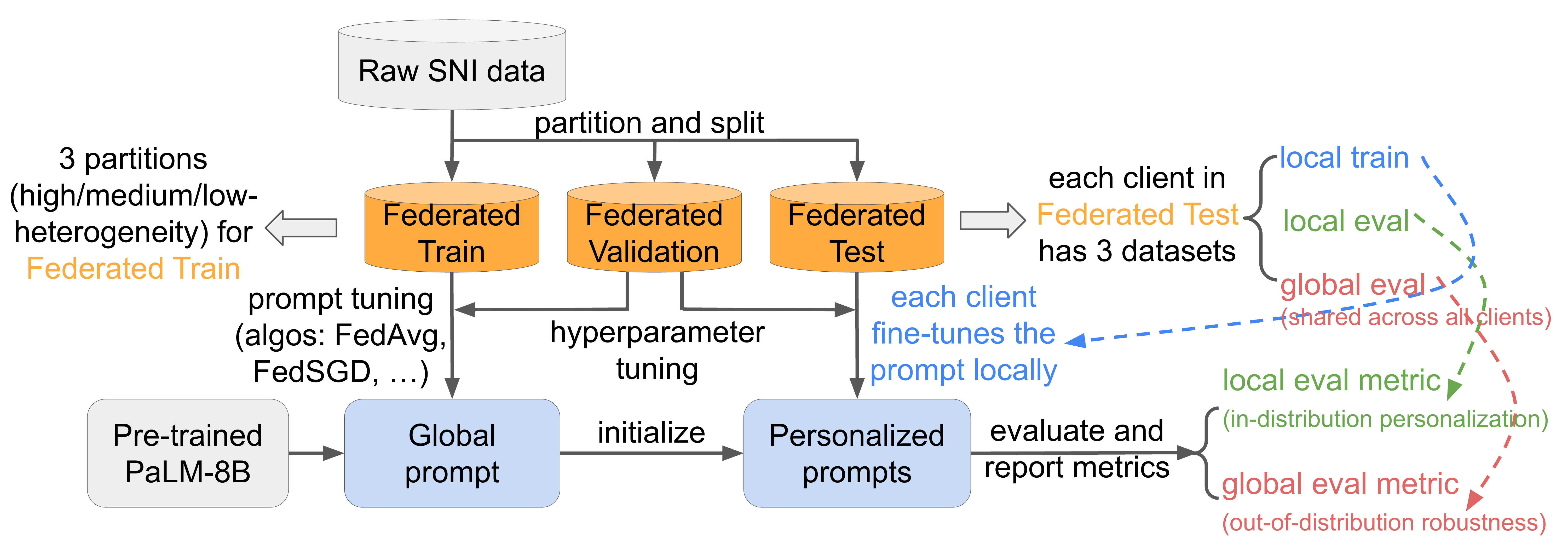}
    \caption{\textbf{Overview of our experimental setup.} We partition and split the raw SNI dataset into three federated datasets: train (used for training a global prompt), validation (used for hyperparameter tuning), and test (used for learning and evaluating the personalized prompts). We experiment with three versions (high/medium/low heterogeneity) of training data. In the test data, each client has three local datasets: a local train set (used for locally fine-tuning the global prompt to produce the personalized prompt) and local and global eval sets (used for evaluating the personalized prompt over the local and global distributions, respectively). The global eval set is shared across all clients, and is formed by sampling from all test clients' local eval sets. See Section~\ref{sec:experiment_setup} for more details.}
    \label{fig:experimental_procedure}
\end{figure}

In this section we detail the framework we use to empirically evaluate 
federated-trained prompts.

\textbf{Datasets.} We construct three federated datasets from Super-NaturalInstructions (SNI) \citep{wang2022super}. SNI is a collection of 1761 diverse NLP tasks belonging to one of 76 {\em task types}. Task types include both text classification and generation types, with Translation, Question Answering, and Question Generation being the most popular. Tasks have on average $\sim$3000 (query, target) pairs, called instances. 
For example, the instances of a  translation task may be pairs of English to French translations of text snippets from a particular corpus. 
This hierarchical structure (task type $\rightarrow$ task $\rightarrow$ instance) allows for partitioning the instances into clients with controllable heterogeneity level.


\begin{table}
  \caption{\textbf{Dataset statistics.}  All partitions have {3520 training clients} and all federated experiments sample {32 training clients/round}. Per-client statistics are the mean total instances and unique tasks and task types found in each training client's dataset (rounded to the nearest integer) $\pm$ standard deviation across training clients. 
  There are 326 test and validation clients each (the same for all partitions), and each test and validation client has approximately 1200 instances.
  }
  \label{table:data}
  \centering
  \begin{tabular}{cccccc}
     \toprule 
    Partition  & \# train clients & Clients/round  & Instances/client    & Tasks/client &  Task types/client \\
    \midrule
    HHF-SNI  & 3520 & 32  & 1201 $\pm$ 17.6 & 1 $\pm$  0.8  & 1 $\pm$  0 \\
    MHF-SNI  & 3520 & 32    & 1201 $\pm$  17.6  &  118 $\pm$  111.2  &  1 $\pm$  0 \\
    LHF-SNI  & 3520 & 32   & 1201 $\pm$ 0.4  &  640 $\pm$  10.8    & 50 $\pm$  1.8  \\
    \bottomrule
  \end{tabular}
\end{table}

\begin{figure} 
  \centering
  \includegraphics[width=0.65\textwidth]{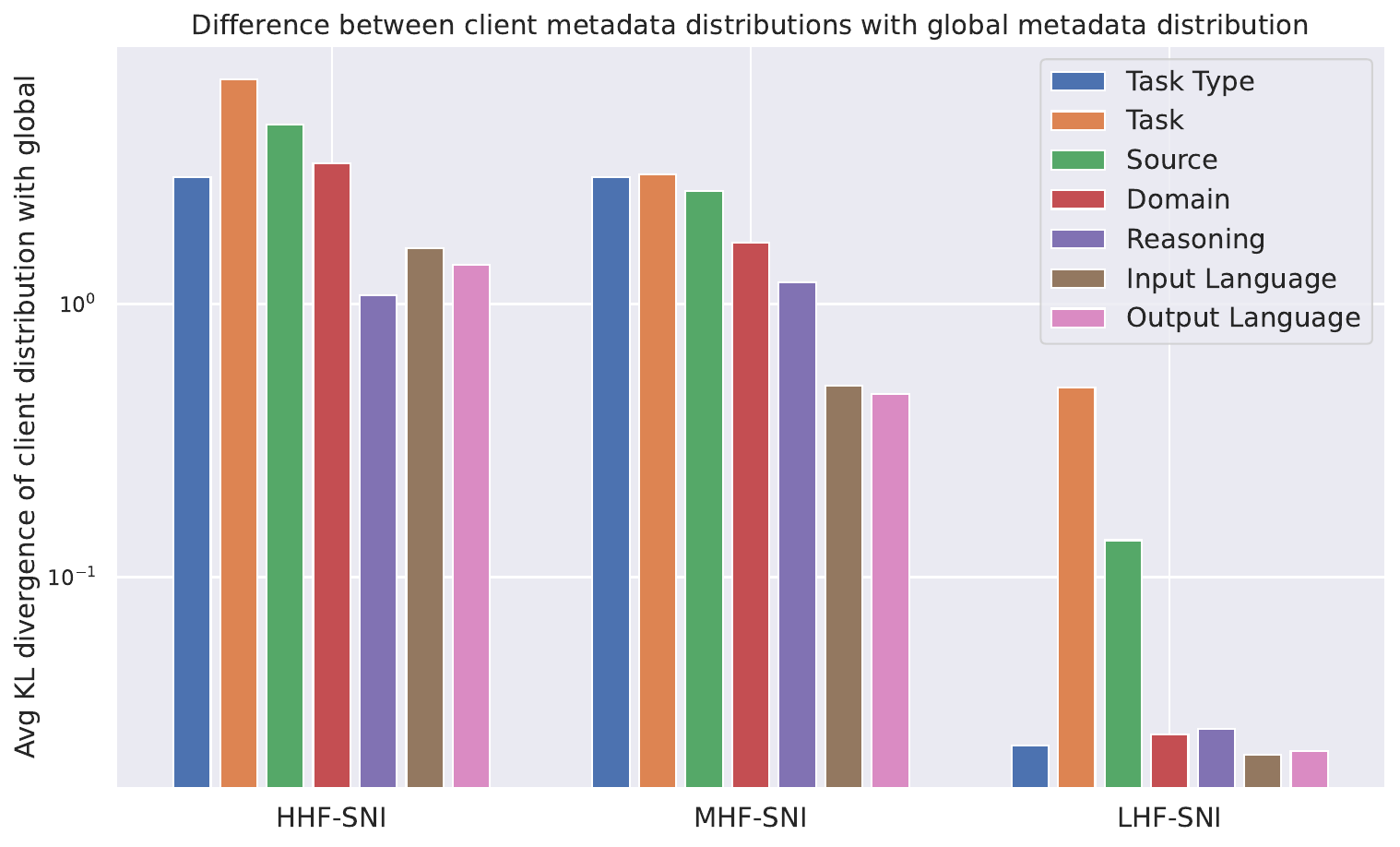}
  \caption{For each of the three training dataset partitions (HHF-SNI, MHF-SNI, LHF-SNI) and each metadata category (Task Type, Task, Source, Domain, Reasoning, Input Language, and Output Language), we plot the average across clients of the  Kullback-Leibler (KL) divergences between the client's metadata category distribution and the global metadata category distribution, in log scale. Among a variety of meta-data categories, clients on average have distributions of this meta-data category that differ from the global distribution to an extent that we would expect from high, medium and low-heterogeneity partitions (the larger the heterogeneity, the greater the difference between client and global distributions).}\label{fig:data_heterogeneity}
\end{figure}

We partition the instances into clients by first splitting them into training, validation, and test sets according to task type. 
We randomly select 7 task types each for testing and validation\footnote{The test task types are Irony Detection, Text Completion, Explanation,
 Overlap Extraction, Question Generation, Dialogue Act Recognition, and Gender Classification.}. Then, we partition the test and validation data into clients by ordering the instances in each task type by task, then breaking these lists into evenly-sized chunks of adjacent instances and designating each chunk to a client. As a result, each client's instances belong to a single task type, and typically a single task. 
Next, we construct three distinct partitions of the training data.
First, we construct a {\em high heterogeneity} partition in exactly the same manner as we partition the validation and test data. We do the same for a {\em medium heterogeneity} partition, except that we shuffle the instances within each task type before dividing them into client chunks, so that each client may have instances from many tasks of the same type. Lastly, we construct a {\em low heterogeneity} partition by shuffling the entire dataset before dividing it into client chunks, thus each client has instances from many tasks of many types. 
All of each training clients' instances are used in federated training, and the same validation and test sets are used for all three partitions. We call these three partitions High Heterogeneity Federated SNI \textbf{(HHF-SNI)}, Medium 
HF-SNI
\textbf{(MHF-SNI)}, and Low 
HF-SNI
\textbf{(LHF-SNI)}, respectively, and provide dataset statistics that verify heterogeneity levels in Table \ref{table:data} and  Figure \ref{fig:data_heterogeneity}.  

\textbf{Model and metric.}
We use the 8 billion-parameter version of the original PaLM \citep{chowdhery2022palm}, which  was trained on  780 billion tokens from sources including social media and Wikipedia\footnote{We choose this model to minimize data leakage, since it was released prior to the release of SNI. Nevertheless, there could still be overlap between its training data and the sources used by SNI.}.
Following \cite{wang2022super}, we use ROUGE-L \cite{lin-2004-rouge} to measure similarity between predicted and target sequences,
with scores in [0, 1] and larger scores indicating greater similarity.

\textbf{Experimental procedure.} We execute a two-stage experimental procedure. In Stage 1, we run federated learning on the training clients to learn global prompt parameters (see Appendix \ref{app:background} for more details on prompt tuning).
In Stage 2, we evaluate the quality of these global parameters by using them to initialize local training (personalization) on each test client. In particular, each test client independently trains a soft prompt on their training set starting from the federated-trained global prompt. As this local training progresses we record the prompt's scores on the corresponding client's test data and on a global test dataset compiled across all of the test clients' test datasets. The local scores serve as the personalization metric, while the global scores serve as the robustness metric. We  hyperparameter tune in Stage 1  by evaluating the global prompt on a global validation dataset collected from all the validation clients, and in Stage 2  by running personalization on the validation clients.  Figure \ref{fig:experimental_procedure} depicts this procedure in detail.

\textbf{Baselines and hyperparameters.}
We study a generalized version of FedAvg proposed in \cite{reddi2021adaptive} that allows for adaptive server and client optimizers\footnoteref{stateless}.
As in \cite{reddi2021adaptive}, we find that using an adaptive server optimizer, in our case Adam, improves over SGD, so all our experiments use Adam on the server side. For the client optimizer\footnoteref{stateless}, we experiment with both Adam and SGD, referring to these versions of FedAvg as \textbf{FedAvg(Adam)} and  \textbf{FedAvg(SGD)}, respectively. Both algorithms make 16 local updates with batch size 32 on 32 sampled clients per round for 300 rounds, and the Adam optimizer is re-initialized from scratch at the start of each selected client's local training round. We also consider \textbf{FedSGD}, in which 32 clients per round send the gradient of the global prompt estimated on 32  instances directly back to the server, and the server updates the global model using Adam. We execute FedSGD for 4800 rounds so that FedSGD processes the same total number of instances as the FedAvg methods. In Appendix \ref{app:experiments}, we explore a version of FedSGD that multiplies the batch size
(rather than the number of communication rounds)  by 16  in order to see the same number of instances as FedAvg, noting that this gave significantly worse results. We also run \textbf{Centralized} training with Adam and batch size 1024 (same effective batch size as FedSGD) for 4800 rounds.

All algorithms optimize prompts of length 10 (tuned in $\{5, 10, 20\}$) with embedding dimension 4096. We tune learning rates, the Adam epsilon parameter, and the weight decay parameter during federated training.
For personalization, we run Adam and tune its learning rate based on the number of epochs available.
We evaluate on 32 test clients, each with training and test sets of 256 and 128 instances, respectively, and a global test set of 2048 instances. Additional details are provided in Appendix \ref{app:experiments}.

\section{Results} \label{sec:experiments}
%
Next, we share personalization (i.e., the local score obtained by evaluating a client's personalized model on this client's local data) vs robustness (i.e., the global score obtained by evaluating the same personalized model over the global test set) curves during personalization. Each point in each plot is the mean (local score, global score) across clients during a personalization epoch,  averaged over two-end-to-end trials with distinct random seeds\footnote{Our observations are consistent across random seeds; see results for individual seeds in Appendix \ref{app:experiments}.}.
These results admit a number of observations. Additional results and details are provided in Appendix \ref{app:experiments}.

\begin{figure}
    \centering
    \begin{subfigure}[b]{0.34\textwidth}
        \centering
        \includegraphics[width=\textwidth]{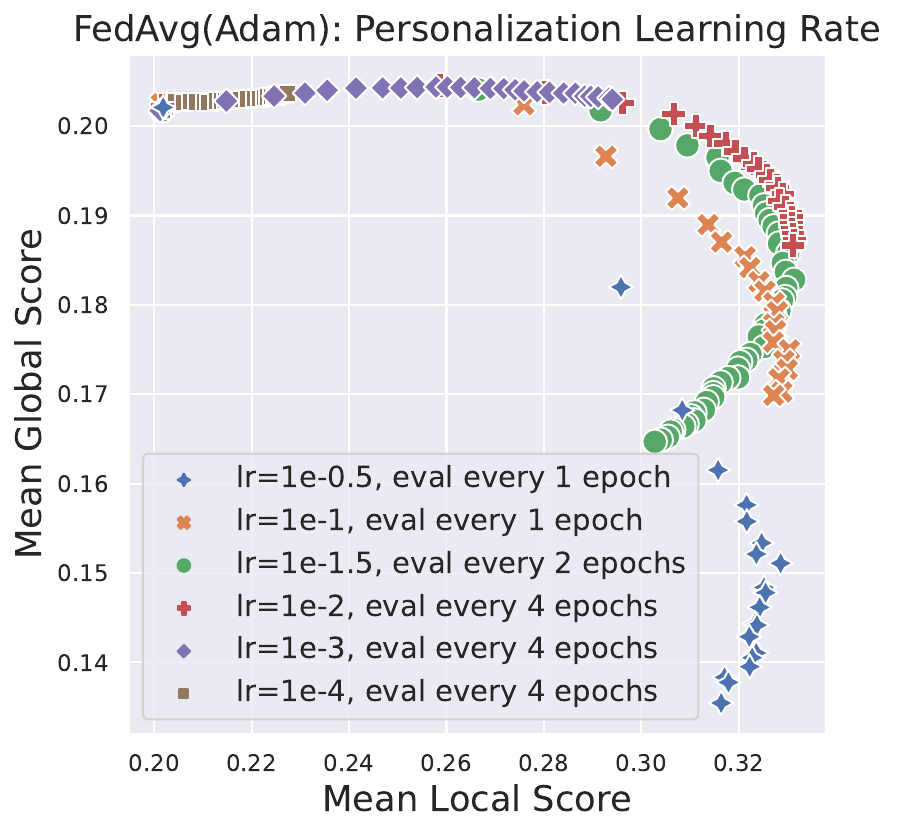}  
    \end{subfigure}
    \hfill
    \begin{subfigure}[b]{0.32\textwidth}
        \centering
        \includegraphics[width=\textwidth]{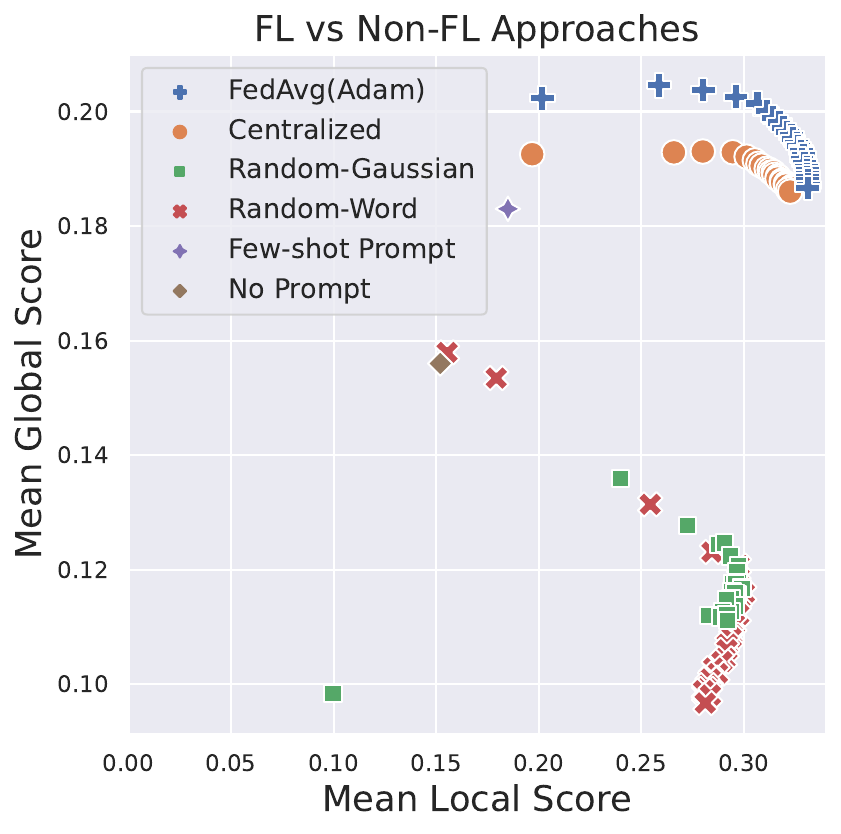}
    \end{subfigure}  
    \hfill
    \begin{subfigure}[b]{0.32\textwidth}
        \centering
        \includegraphics[width=\textwidth]{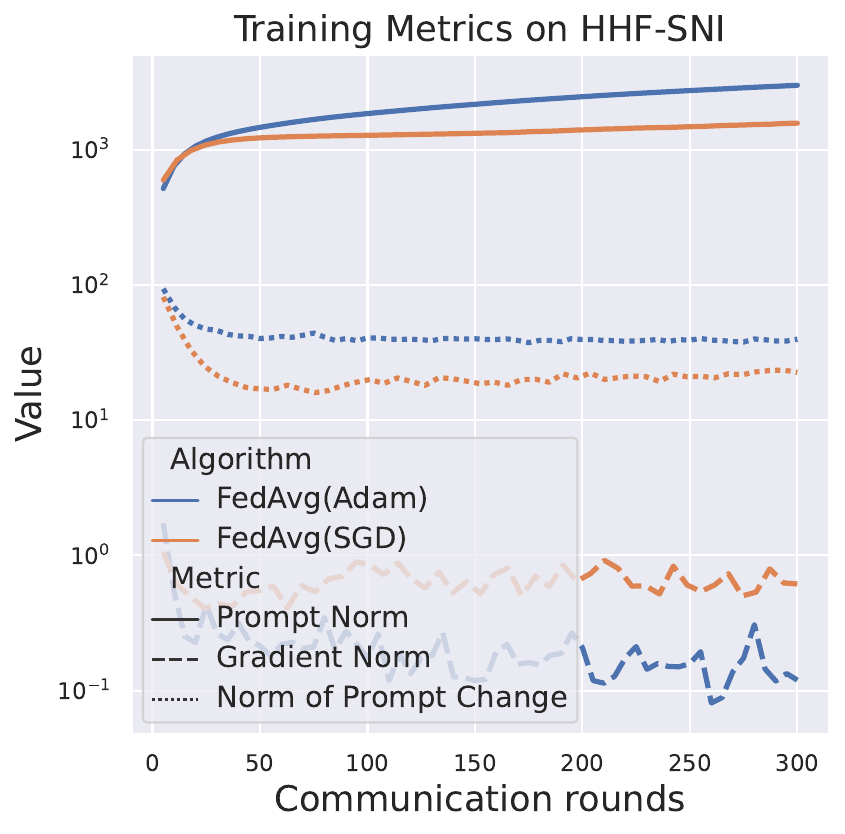}
    \end{subfigure}  
    \caption{ \textbf{(Left)} Global and local scores during personalization with varying learning rates from a prompt trained on HHF-SNI by FedAvg(Adam). All runs besides those with the largest two learning rates are run for 100 epochs, and otherwise 20 epochs. \textbf{(Center)} Global and local scores during 100 epochs (high computation) of personalization starting from FedAvg(Adam) and Centralized-pre-trained prompts and random initializations (with evaluations every 4 epochs), plus global and local scores with no prompt and few-shot (engineered) prompts.
    \textbf{(Right)} Global prompt norm, average gradient norm across clients, and norm of prompt change on consecutive rounds during FedAvg(Adam) and FedAvg(SGD) training. 
    All norms are Frobenius.
    }
    \label{fig:nonfl}
\end{figure}

\textbf{Observation 1: Choice of personalization learning rate induces computation vs robustness trade-off.} Figure \ref{fig:nonfl}(Left) plots global and local scores during personaliztion with varying learning rates starting from a prompt pre-trained on HHF-SNI with FedAvg(Adam). These results show that the personalization vs robustness trade-off is heavily dependent on the personalization learning rate. In particular, higher global scores can be maintained by personalizing with smaller learning rates, but at the cost of requiring more epochs to reach the maximal local scores. Specifically, with learning rate $10^{-0.5}$, the average local score reaches 0.32 within 10 epochs and the average global score drops to $~0.15$, and with learning rate $10^{-2}$, 64 epochs are required to reach average local score 0.32, but the average global score does not drop below 0.19. In effect, this induces a computation vs robustness trade-off: more robustness necessitates more computation.

 This motivates us to consider two distinct regimes for personalization: (1) \textbf{High Computation}, in which each client executes 100 epochs of personalization, and (2) \textbf{Low Computation}, in which each client executes 10 epochs of personalization, with learning rates tuned to achieve the best local score (0.32) with minimal drop in global score for each regime. We use regime (1) to compare different pre-training algorithms, as this allows the best performance for each algorithm (Observations 2 and 3). Then, we conclude by showing the more severe forgetting  in regime (2) can be mitigated by incorporating a number of heuristics (Observation 4).


\textbf{Observation 2: Benefit of FL pre-training.} Figure \ref{fig:nonfl}(Center) considers the High Computation regime and shows global vs local score curves for prompts pre-trained with FedAvg(Adam) and centralized training, along with prompts initialized by sampling from a Gaussian distribution (``Random Gaussian'') and by sampling 10 token embeddings from the PaLM token embedding matrix (``Random Word'') \cite{gu2021ppt}. FedAvg(Adam) yields the best personalization vs robustness trade-off, especially compared to the random initializations. Surprisingly, FedAvg(Adam) outperforms centralized training, although centralized training achieves smaller training loss (see Appendix \ref{app:experiments}), as expected due to possible objective inconsistency for FedAvg \cite{wang2020tackling}. 
FedAvg(Adam) also outperforms both No Prompt and Few-shot Prompts, which are constructed using instructional examples according to the best procedure reported in \cite{wang2022super}; please see Appendix \ref{app:experiments} for details.

\begin{figure}
    \centering
    \begin{subfigure}[b]{0.32\textwidth}
        \centering
        \includegraphics[width=\textwidth]{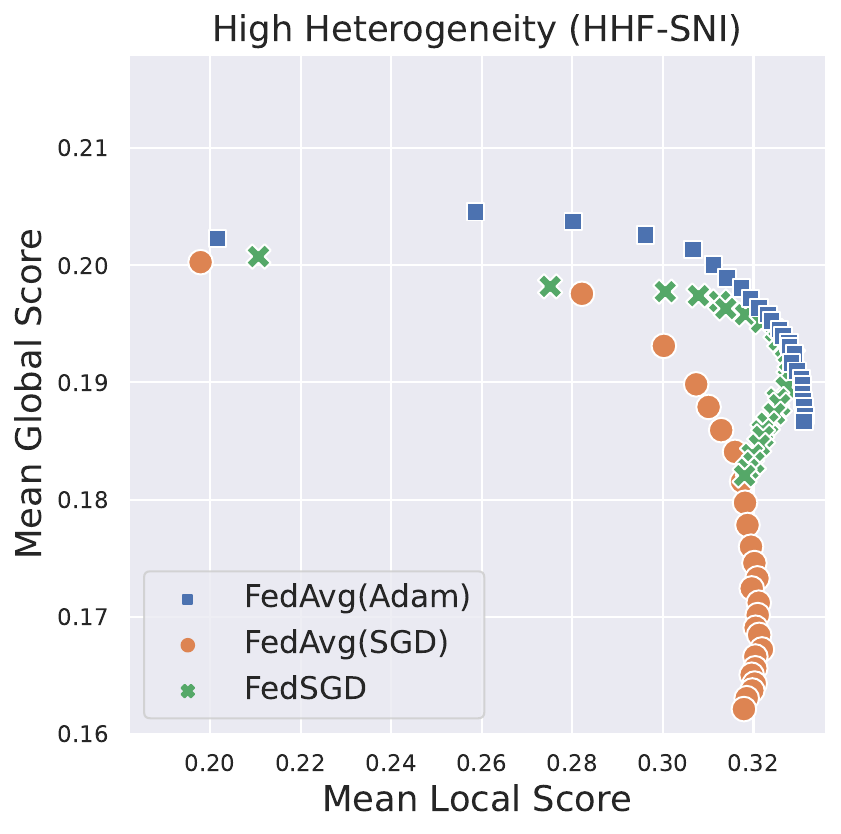}  
    \end{subfigure}
    \hfill
    \begin{subfigure}[b]{0.32\textwidth}
        \centering
        \includegraphics[width=\textwidth]{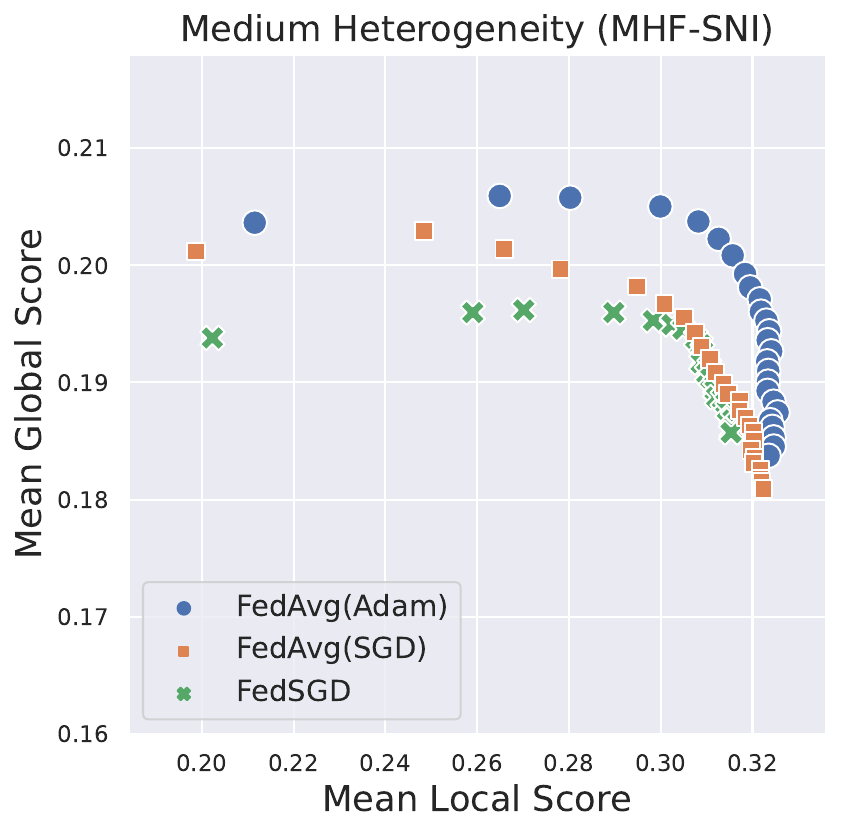}
    \end{subfigure}  
    \hfill
    \begin{subfigure}[b]{0.32\textwidth}
        \centering
       \includegraphics[width=\textwidth]{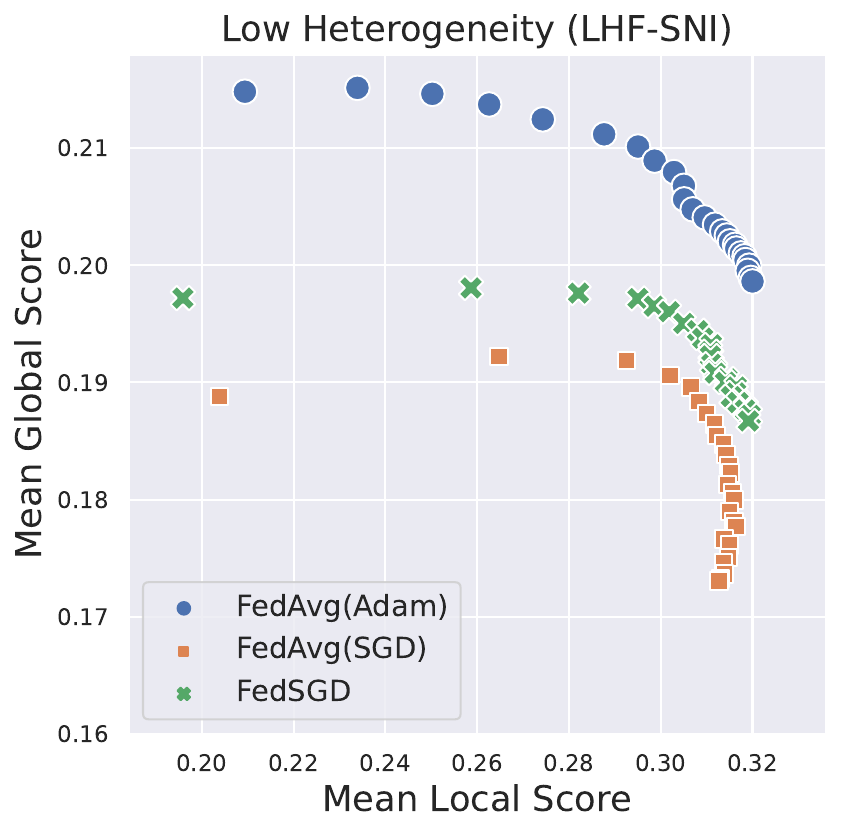}
    \end{subfigure}  
    \caption{\textbf{High Computation} regime: scores evaluated every 4 epochs during \textbf{100 epochs} of  personalization starting from prompts pre-trained  by FedAvg(Adam), FedAvg(SGD) and FedSGD on 
    \textbf{(Left)} HHF-SNI,   \textbf{(Center)} MHF-SNI, and  \textbf{(Right)} LHF-SNI.
    }
    \label{fig:fl}
\end{figure}

\textbf{Observation 3a: Importance of adaptive client optimizer\footnoteref{stateless}.} Figure \ref{fig:fl} compares prompts trained with FedAvg(Adam), FedAvg(SGD), and FedSGD during personalization in the High Computation regime. FedAvg(Adam) outperforms FedAvg(SGD) on all three training partitions, highlighting the benefit of using an adaptive client optimizer\footnote{Often,  the client optimizer in FL is SGD, motivated by the added memory cost of Adam \cite{reddi2021adaptive}. However, this cost is linear in the number of trainable parameters, so it is small for prompt tuning -- recall we use a prompt length of 10 tokens with embedding dimension 4096, for a total of only 40960 trainable parameters.
}.
It is well-known that adaptive optimization enhances full-model transformer training \cite{zhang2020adaptive,kunstner2023noise,pan2023toward}, but to our knowledge this has not yet been observed for prompt tuning.
Based on Figure \ref{fig:nonfl}, we conjecture that Adam's benefit stems from prompt tuning's flat loss landscape relative to prompt scale. For both FedAvg(Adam) and FedAvg(SGD), gradient norms are three orders of magnitude smaller than prompt norms throughout training. This means that the SGD updates are relatively insignificant, unlike the Adam updates that have normalized gradient and a momentum term that scales with the prompt norm.
Thus, FedAvg(SGD) has smaller prompt changes than FedAvg(Adam),  despite having a client learning rate 100x larger (see Table \ref{table:lr}). 











\textbf{Observation 3b: Importance of multiple local updates.} Figure \ref{fig:fl} also shows that FedAvg(Adam) outperforms FedSGD, especially with lower training data heterogeneity. Multiple recent works have  noticed the superiority of FedAvg-trained models as initializations for personalization compared to FedSGD-trained models \cite{charles2023towards,collins2022fedavg,jiang2019improving}, but these works did not consider the robustness to forgetting after personalization (nor prompt tuning). In contrast, here we observe that the improvement due to FedAvg is mostly due to higher {\em  global scores}. 
Since we use Adam as the server optimizer for FedSGD, the improvement of FedAvg(Adam) cannot be due to its updates being adaptive, but must be due to making {multiple} of them between communication. 













\begin{figure}
    \centering
    \begin{subfigure}[b]{0.32\textwidth}
        \centering
        \includegraphics[width=\textwidth]{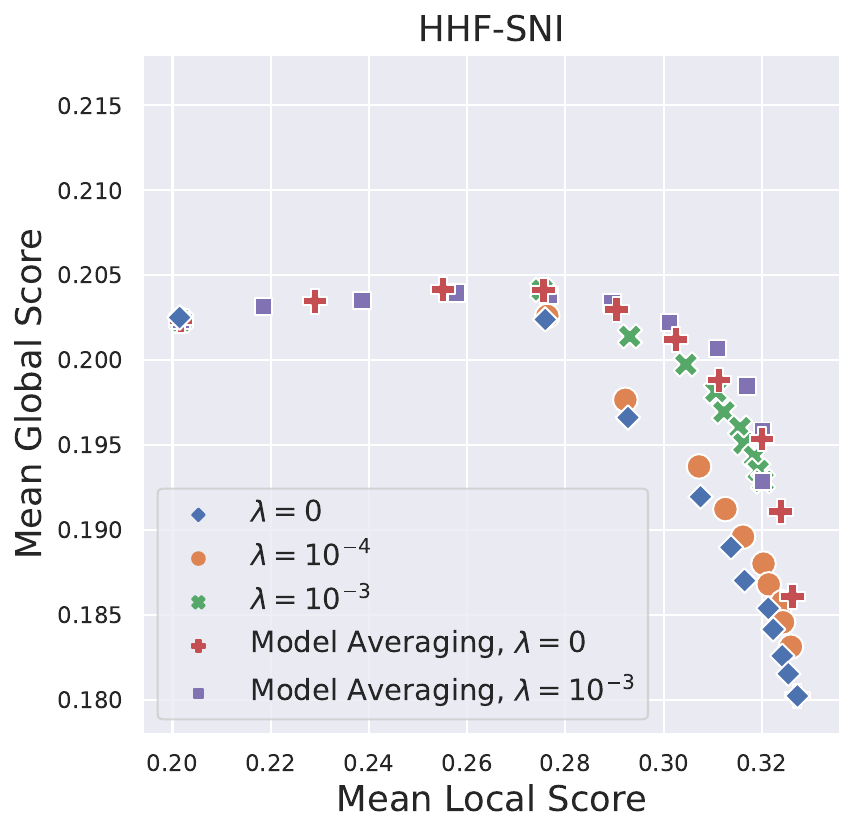}  
    \end{subfigure}
    \hfill
    \begin{subfigure}[b]{0.32\textwidth}
        \centering
        \includegraphics[width=\textwidth]{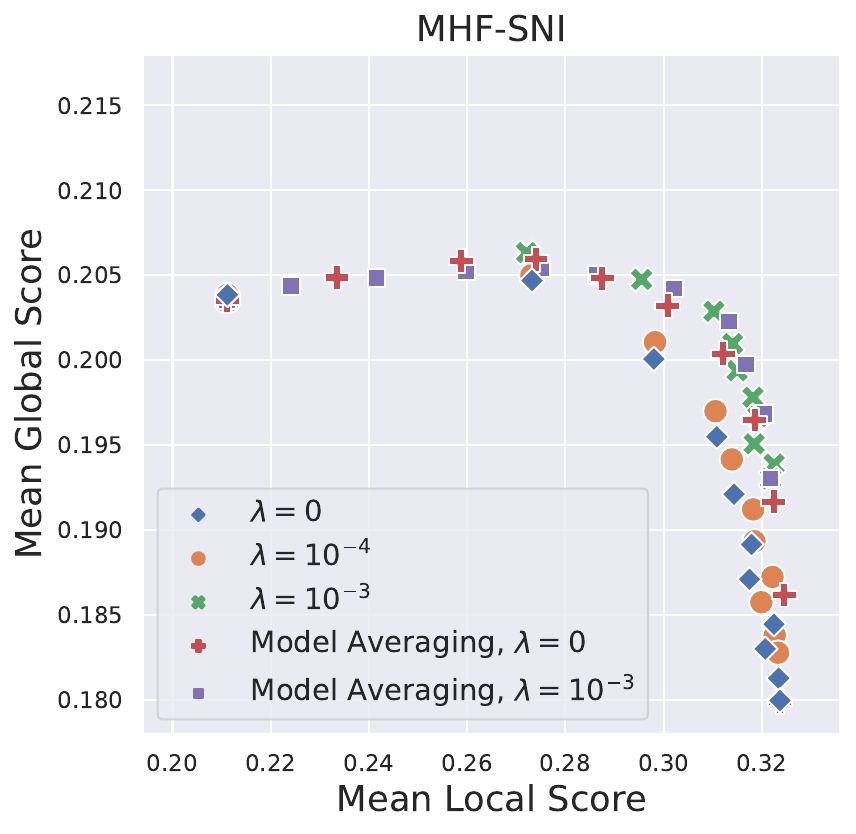}
    \end{subfigure}  
    \hfill
    \begin{subfigure}[b]{0.32\textwidth}
        \centering
        \includegraphics[width=\textwidth]{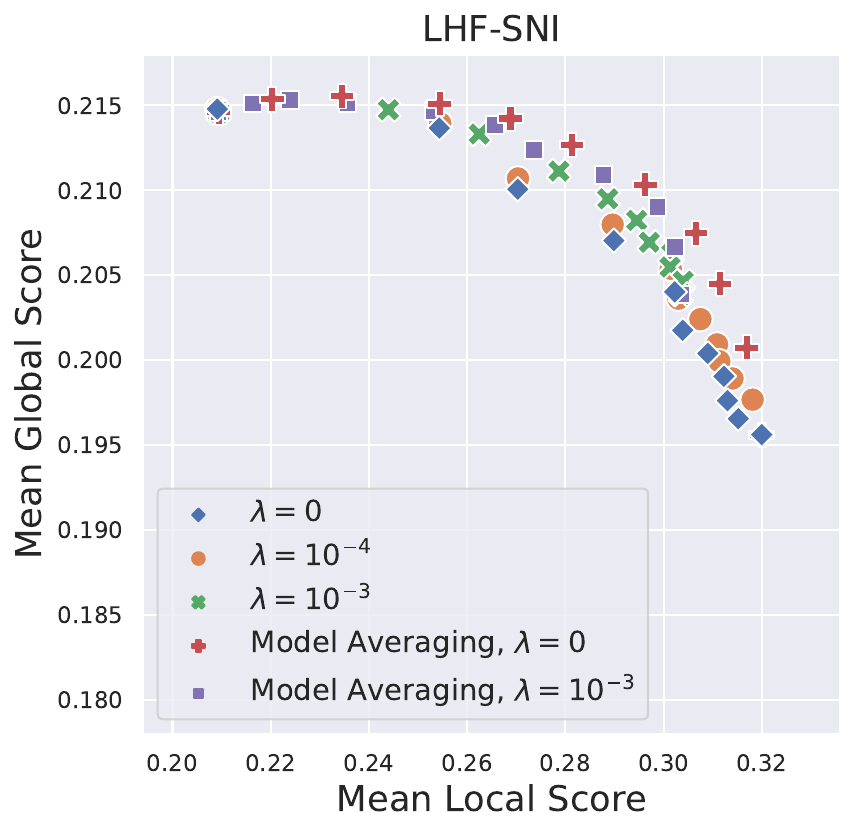}
    \end{subfigure}  
    \caption{
    \textbf{Low Computation} regime: scores evaluated every epoch during \textbf{10 epochs} of personalization with robust-l2 regularization with parameter $\lambda$, and possibly model averaging, starting from prompts trained by FedAvg(Adam) on
    \textbf{(Left)} HHF-SNI,   \textbf{(Center)} MHF-SNI, and  \textbf{(Right)} LHF-SNI.
    }
    \label{fig:heuristics}
\end{figure}

\textbf{Observation 4: Personalization-robustness trade-off can be improved by personalization heuristics.}
Figure \ref{fig:heuristics} considers the Low Computation regime, in which each test client only executes 10 personalization epochs on its training set of 256 instances. Here, we evaluate two heuristics to improve the personalization vs robustness trade-off: (1) $\ell_2$ regularization and (2) model averaging~\cite{wortsman2022model, wortsman2022robust, ilharco2022patching}. 
Let $P_{\text{glob}}$ be the global prompt resulting from federated training, $P_{i}$ be the prompt the Client $i$ personalizes, and $P_{i, 10}$ be the prompt resulting from $10$ epochs of personalization to Client $i$.
For (1), we add $\ell_2$ regularization with parameter $\lambda$ to the loss that penalizes the distance of the personalized prompt from the global prompt, specifically $\frac{\lambda}{2} \|P_{i} - P_{\text{glob}}\|_F^2$ is added to the loss. For (2), we first run full personalization, then compute final client-specific prompts by interpolating the global and personalized prompts as
$$ P_{i,10,\alpha} = \alpha P_{i, 10} + (1 - \alpha) P_{\text{glob}}$$
for $\alpha \in \{0, 0.1,0.2, ..., 1\}$. Each point for ``Model Averaging'' in each plot in Figure \ref{fig:heuristics} corresponds to the mean (local, global) scores across client prompts $\{P_{i,10,\alpha}\}_i$ for a particular value of $\alpha$,  with increasing $\alpha$ moving from left to right in each plot. 
Figure \ref{fig:heuristics} shows that both $\ell_2$ regularization and model averaging, as well as their combination (``Model Averaging, $\lambda = 10^{-3}$''), improve the personalization-robustness trade-off for FedAvg(Adam)-trained prompts.

\section{Conclusion}

Our benchmarking experiments evince the effectiveness of FL as a prompt pre-training mechanism, the importance of using adaptive optimizers for prompt tuning, and the superiority of FedAvg over FedSGD for prompt tuning. We also demonstrate that simple approaches such as reducing the personalization learning rate, adding $\ell_2$ regularization, and model averaging can improve the personalization vs robustness trade-off for federated-trained prompts.
Important open questions stemming from our results include why  Adam is superior to SGD for the client optimizer for prompt tuning and why making multiple local updates per round (i.e. FedAvg-style training) leads to more generalizable prompts then making only one update per round (FedSGD), even though both methods see the same total amount of data.
Furthermore, our work is limited to exploring simple FL algorithms, without privacy guarantees, on a single model (PaLM-8B). 
Studying how the personalization-robustness trade-off can be further improved by leveraging more sophisticated personalized FL algorithms, how model size affects the personalization-robustness trade-off, and how to characterize the privacy-utility trade-off for federated prompt tuning also remains as essential future work.

\section*{Acknowledgements}
We are grateful to Sean Augenstein, Eugene Bagdasaryan, Zachary Charles, Krishna Pillutla and Zheng Xu for helpful discussions and comments. 

\bibliography{ref.bib}
\bibliographystyle{plainnat}
\appendix
\section{Formal Problem Setup} \label{app:background}
\textbf{Federated prompt tuning.} We consider a federated learning scenario consisting of $n$ clients that communicate with a central server. For every $i\in [n]$, Client $i$ has a dataset $\mathcal{D}_i\coloneqq \{(x_{i,j}, y_{i,j})\}_{j=1}^{m_i}$ consisting of $m_i$ query-target pairs $(x_{i,j}, y_{i,j})$, where each query $x_{i,j}$ and target $y_{i,j}$ is a variable-length text sequence. All clients also have a copy of a language model with parameters $\theta$, a tokenizer $\tau$ mapping text to a list of one-hot encodings of tokens, and a token embedding matrix $E \in \mathbb{R}^{e \times v}$, where $e$ is the embedding dimension and $v$ is the vocabulary size.

When provided an input $x$, the language model computes the conditional distribution of tokenized targets given the embedding of the tokenized input query, namely  $\mathbb{P}_{\theta}(\tau(Y)|E \tau(x))$, in order to generate text predictions. A natural idea to more accurately estimate the conditional distribution of $\tau(Y)$ is to add text (a prompt) $p$ to the input query that provides information about the relationship between inputs and targets for each task at hand, such as instructions or examples of gold-standard $(x,y)$ pairs. In other words, the idea is that $\mathbb{P}_{\theta}(\tau(Y)|E \tau([p,x])) \equiv \mathbb{P}_{\theta}(\tau(Y)|[E\tau(p), E \tau(x)]) $ should be a more accurate estimation of the true conditional distribution of $Y$ given $x$ for carefully chosen $p$. This approach is known as  
{\em in-context learning} or {\em prompt engineering} 
%
and has led to many successful adaptations of LLMs \citep{brown2020language}.
However, these discrete text prompts cannot  be easily  optimized, and
restricting the embedded prompt $E\tau(p)$ to columns in $E$ limits the information it can convey about the relationship between $Y$ and $X$.


{\em Prompt tuning} \citep{lester2021power} addresses these concerns by optimizing a ``soft'' prompt in embedding space. For some number of tokens $k$, prompt tuning aims to learn a matrix $P \in \mathbb{R}^{e \times k}$ that conditions the model for more accurate predictions when prepended to the {\em embedding} of the input text tokens, i.e. the new model is given by 
$\mathbb{P}_{\theta}(\tau(Y)|[P, E \tau(x)])$. In this case, the gradient of the loss of $\mathbb{P}_{\theta}(\tau(Y)|[P, E \tau(x)])$ with respect to $P$ can be easily computed via backpropagation, and we can optimize $P$ with standard gradient-based methods. This loss is the cross-entropy loss, in particular, the loss as a function of $P$ for Client $i$ in our federated setting is:
% ...
\begin{align}
    \mathcal{L}_i(P)\coloneqq -\frac{1}{m_i} \sum_{j=1}^{m_i} \log(\mathbb{P}_{\theta}(\tau(y_{i,j})|[P,E \tau(x_{i,j})]))
\end{align}
During federated training, the server aims to minimize the average loss across clients, namely
$\mathcal{L}(P) \coloneqq \frac{1}{n}\sum_{i=1}^n \mathcal{L}_i(P)$,
{\color{black}
and towards this end can apply standard Federated Learning algorithms such as FedAvg and FedSGD. Importantly, only the prompt embedding matrix $P$ must be communicated between server and clients, as depicted in Figure \ref{fig:fed_prompt_tuning}.
}





\textbf{Personalization and robustness.} Due to the heterogeneity of the client datasets $\mathcal{D}_1, \dots, \mathcal{D}_n$, the global prompt $P_{\text{glob}}$ found by running federated learning on $\mathcal{L}(P)$ may not perform well on each client's local data. This can be addressed by personalizing $P_{\text{glob}}$ to each client. Formally, we consider a new set of $n_{\text{test}}$ clients with datasets $\mathcal{D}_{n+1},\dots,\mathcal{D}_{n + n_{\text{test}}}$ that are split into training and test sets, i.e. $\mathcal{D}_{i} = \mathcal{D}_{i}^{\text{train}}\cup \mathcal{D}_{i}^{\text{test}} $ for all $i=n+1,\dots, n+n_{\text{test}}$. During personalization, Client $i$ updates $P_{\text{glob}}$ using its local training dataset $\mathcal{D}_{i}^{\text{train}}$ to obtain a prompt $P_i$. The level of personalization achieved by this prompt is evaluated using $\mathcal{D}_{i}^{\text{test}}$. However, it is also of interest to know how robust $P_{\text{glob}}$ is to personalization, as we do not want $P_i$ to have forgotten all of the global information it acquired during federated training. So, $P_i$ is also evaluated on a global test dataset compiled across all client test datasets $\mathcal{D}_{n}^{\text{test}}, \dots, \mathcal{D}_{n+n_{\text{test}}}^{\text{test}} $ to obtain a robustness score. These local  personalization and robustness scores are ultimately aggregated across clients and used for final evaluation  of the federated algorithm used to obtain $P_{\text{glob}}$.


\section{Additional Dataset Details}
Of the 76 total task types in SNI, we excluded three types (Punctuation Error Detection, Paper Review, Speaker Relation Classification) because they did not have a sufficient amount of data for one client  and one type, Mathematics, because the PaLM tokenizer cannot properly interpret numerical text input.
The data was split into train/validation/test sets by randomly selecting 10\% of the remaining task types each for validation and testing, and designating the rest for training.
The test task types are [Irony Detection, Mathematics, Text Completion, Explanation, Overlap Extraction, Question Generation, Dialogue Act Recognition, Gender Classification] and the validation types are [Answer Verification, Information Extraction, Dialogue Generation,
 Commonsense Classification, Word Relation Classification, Answerability Classification, Sentence Ordering]. There are 326 total test clients and 326  total validation clients, although we only use 32 test clients, sampled uniformly from the full set of 326 test clients, in our results.



\section{Further Experiments and Details} \label{app:experiments}

{}

{

\textbf{Hyperparameters.} In all training runs, we initialized the prompts by sampling each element i.i.d. from $\mathcal{N}(0, 0.25)$, noting that results from \cite{lester2021power} showed that prompt initialization does not significantly affect performance at the model scale we consider ($\sim10^{10}$ parameters). We tried prompt lengths of 5, 10, and 20, and saw that length 10 generally outperformed length 5, but there was no improvement going from length 10 to length 20,  (see Figure \ref{fig:sgd_p13n}) so we used length 10 for all other runs.
We tuned client and server learning rates in  $\{10^{-2}, 10^{-1}, 10^0, 10^{1}\}$ using the global validation set separately for each algorithm and each of the three training partitions, plus centralized. The resulting learning rates are found in Table \ref{table:lr}. We tuned weight decay parameter in $\{0, 10^{-2}\}$, and Adam epsilon parameter in $\{10^{-8}, 10^{-6}, 10^{-4} \}$ on HHF-SNI and the centralized dataset, and observed that no weight decay and Adam $\epsilon=10^{-8}$ worked best in all cases.
We used $\beta_1 = 0.99$ and $\beta_2=0.999$ for Adam. In each trial, we used the prompt that achieved the highest global validation score during training for personalization.
Regarding model and evaluation parameters, we set the maximum input query length to 1024 tokens and output length to 128 tokens for training and 10 tokens for evaluation, and the decoding temperature to 0, following \cite{wang2022super}. For examples with multiple targets, we take the max score over targets, again following \cite{wang2022super}.


}

\begin{table}
  \caption{\textbf{Training learning rates.} All learning rates were tuned in $\{0.01, 0.1, 1, 10\}$ and chosen based on the global validation score they led to during training. The resulting values are shown here, as (server learning rate, client learning rate) if applicable. Centralized training used Adam with learning rate 1, tuned in the same set.}
  \label{table:lr}
  \centering
  \begin{tabular}{llll}
     \toprule
    Algorithm    & HHF-SNI    & MHF-SNI &  LHF-SNI \\
    \midrule
    FedAvg(Adam) - prompt length 10    & (1, 0.1) & (0.1, 1)   & (0.1, 1)  \\
    FedAvg(SGD)  - prompt length 10     & (1, 10) & (0.1, 10)  & (1, 10)  \\
    FedSGD  - prompt length 10    & 1  &    1 & 1  \\
    FedSGD-LB  - prompt length 10    & 0.01 & 0.1    &0.1  \\
    \bottomrule
  \end{tabular}
\end{table}

\begin{table}
  \caption{\textbf{Adam personalization learning rates.} Personalization learning rates were tuned in $\{10^{-3}, 10^{-2}, 10^{-1.5}, 10^{-1}\}$.}
  \label{table:lr}
  \centering
  \begin{tabular}{llll}
     \toprule
    Algorithm    & HHF-SNI    & MHF-SNI &  LHF-SNI \\
    \midrule
    FedAvg(Adam) - High Computation     & $10^{-2}$ & $10^{-2}$    & $10^{-2}$   \\
    FedAvg(Adam) - Low Computation     & $10^{-1}$ & $10^{-1}$    & $10^{-1}$   \\
    FedAvg(SGD) - High Computation    &  $10^{-2}$ & $10^{-3}$    & $10^{-2}$    \\
    FedAvg(SGD) - Low Computation    &  $10^{-1}$ & $10^{-2}$    & $10^{-1}$    \\
    FedSGD  - High Computation    &   $10^{-1.5}$ & $10^{-2}$    & $10^{-2}$   \\
    FedSGD  - Low Computation    &   $10^{-1}$ & $10^{-1}$    & $10^{-1}$   \\
    FedSGD-LB  - High Computation   &  $10^{-3}$ & $10^{-3}$    & $10^{-3}$ \\
    Centralized  - High Computation   &  $10^{-2}$ & $10^{-2}$    & $10^{-2}$ \\
    Random-Gaussian  - High Computation   &  $10^{-2}$ & $10^{-2}$    & $10^{-2}$ \\
    Random-Word  - High Computation   &  $10^{-2}$ & $10^{-2}$    & $10^{-2}$ \\
    \bottomrule
  \end{tabular}
\end{table}






\subsection{Additional results}

In this section we provide additional empirical results. Unless otherwise noted, all experiments run personalization with Adam on a dataset  of  size 256.

\begin{figure}
    \centering
        \includegraphics[width=0.37\textwidth]{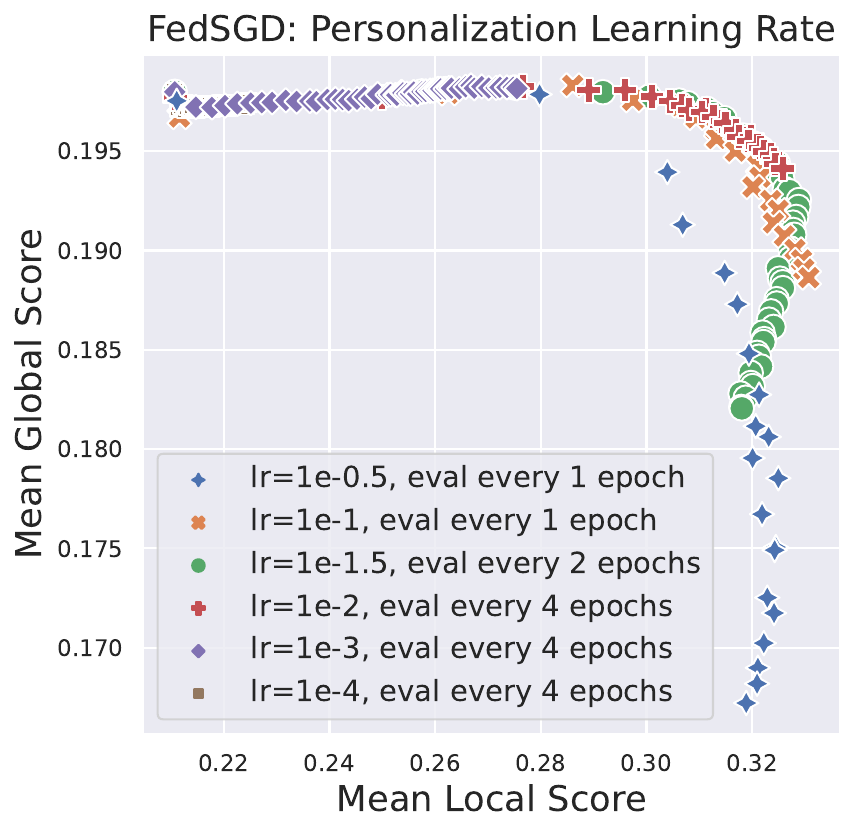}  
    \caption{ Mean global and local scores across test clients during personalization with varying learning rates from a prompt trained on HHF-SNI by \textbf{FedSGD}. All runs besides those with the largest two learning rates are run for 100 epochs, and otherwise 20 epochs.}
    \label{fig:fedsgd-lr}
\end{figure}

\textbf{Role of personalization learning rate with FedSGD-trained prompts.} In Figure \ref{fig:fedsgd-lr} we verify that using a smaller personalization learning rate improves the personalization-robustness trade-off for FedSGD-trained prompts, just like we observed for FedAvg(Adam)-trained prompts in Figure \ref{fig:nonfl}(Left). Again, increased robustness (higher global scores) comes at the cost of additional personalization epochs required to reach high local scores.

\begin{figure}
    \centering
    \begin{subfigure}[b]{0.32\textwidth}
        \centering
        \includegraphics[width=\textwidth]{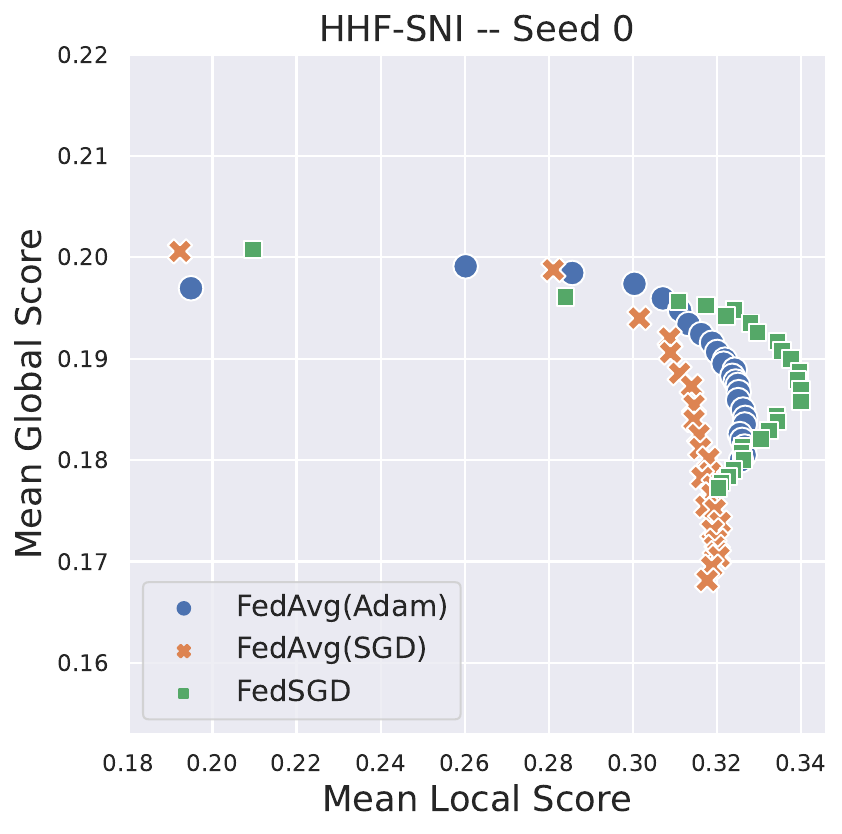}  
    \end{subfigure}
    \hfill
    \begin{subfigure}[b]{0.32\textwidth}
        \centering
        \includegraphics[width=\textwidth]{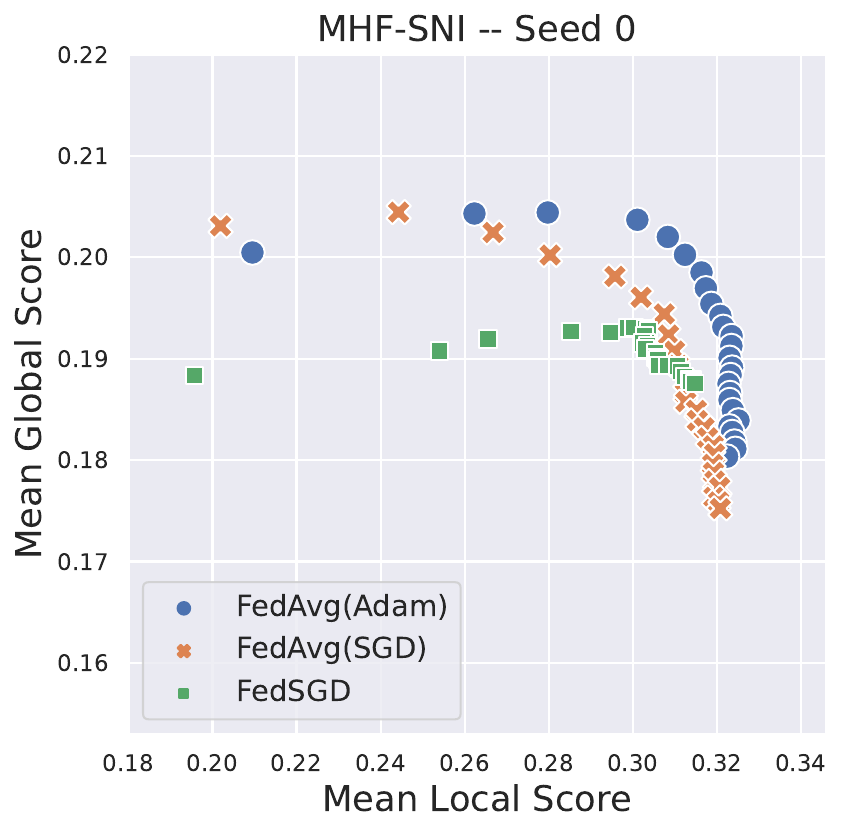}
    \end{subfigure}  
    \hfill
    \begin{subfigure}[b]{0.32\textwidth}
        \centering
        \includegraphics[width=\textwidth]{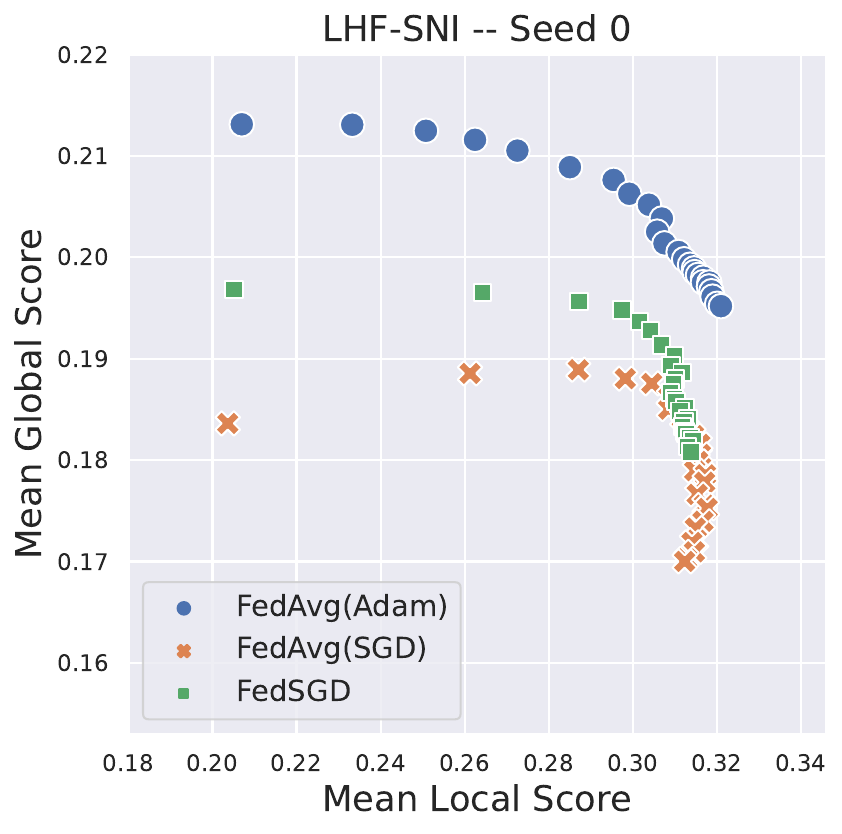}
    \end{subfigure}  
    \caption{\textbf{Version of Figure \ref{fig:fl} with random seed 0.} Mean global and local scores across test clients evaluated every 4 epochs during \textbf{100 epochs} of  personalization (High Computation regime) starting from prompts pre-trained  by FedAvg(Adam), FedAvg(SGD) and FedSGD with random seed 0 on 
    \textbf{(Left)} HHF-SNI,   \textbf{(Center)} MHF-SNI, and  \textbf{(Right)} LHF-SNI.
    }
    \label{fig:seed0}
\end{figure}

\begin{figure}
    \centering
    \begin{subfigure}[b]{0.32\textwidth}
        \centering
        \includegraphics[width=\textwidth]{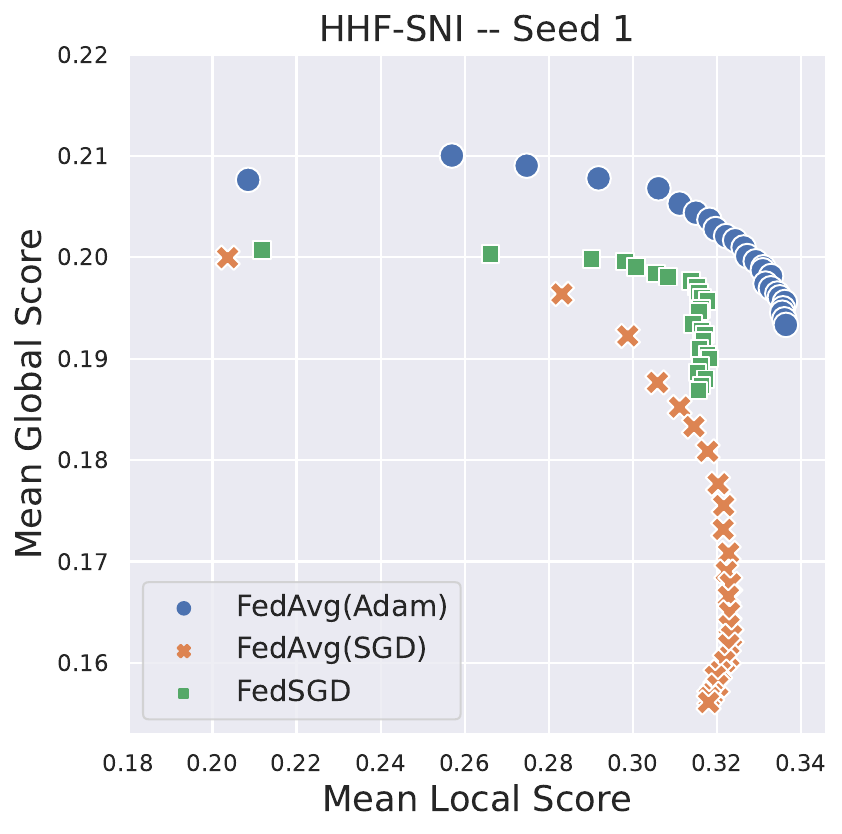}  
    \end{subfigure}
    \hfill
    \begin{subfigure}[b]{0.32\textwidth}
        \centering
        \includegraphics[width=\textwidth]{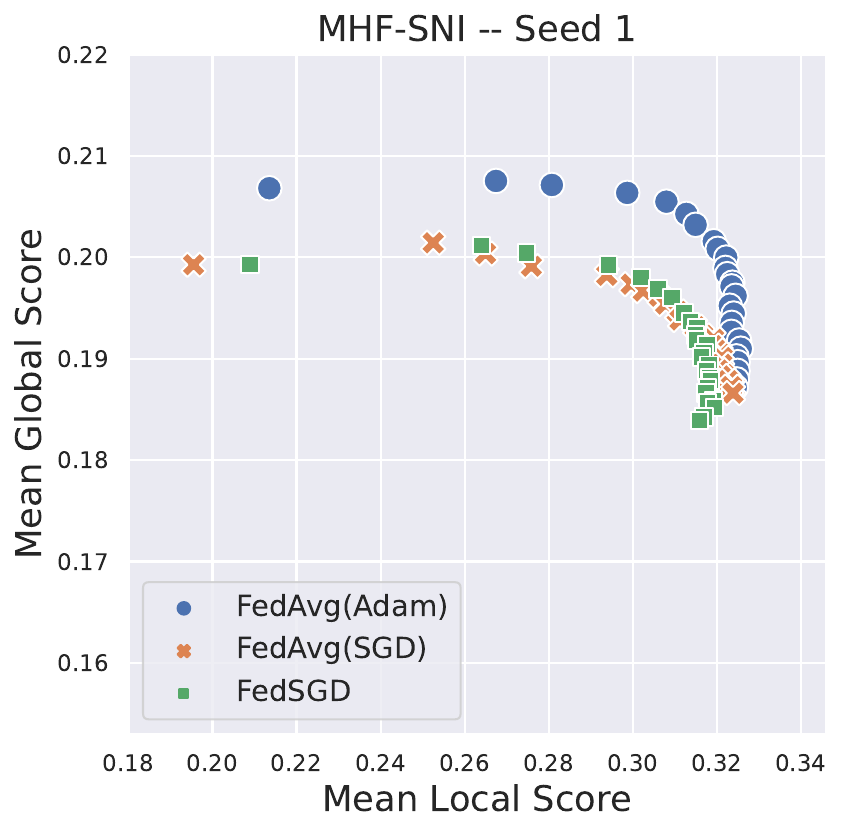}
    \end{subfigure}  
    \hfill
    \begin{subfigure}[b]{0.32\textwidth}
        \centering
        \includegraphics[width=\textwidth]{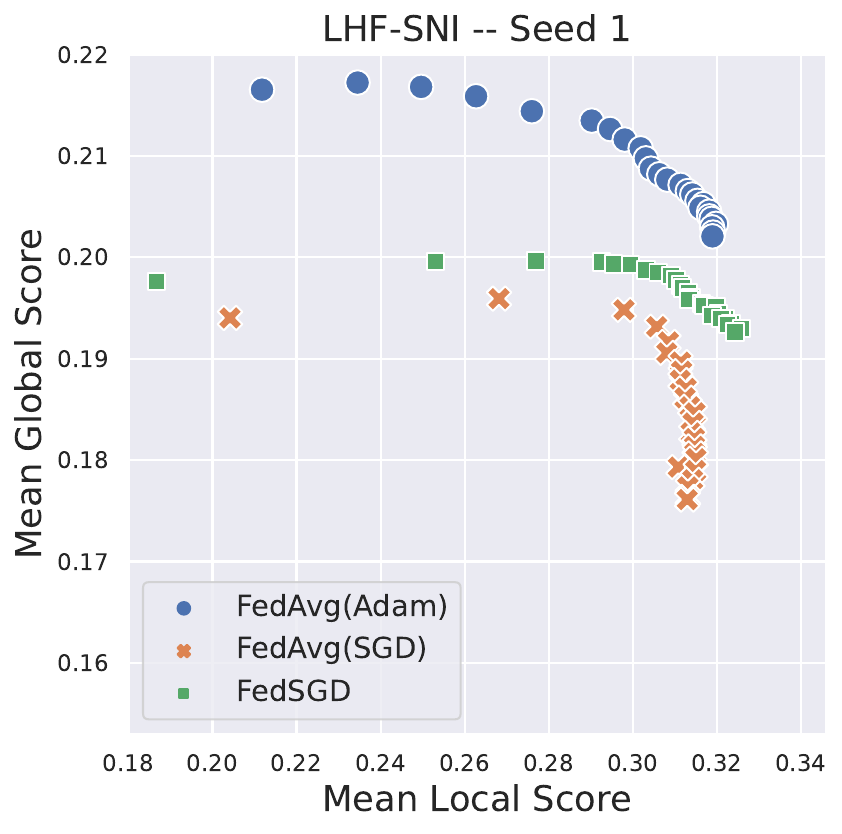}
    \end{subfigure}  
    \caption{\textbf{Version of Figure \ref{fig:fl} with random seed 1.} Mean global and local scores across test clients evaluated every 4 epochs during \textbf{100 epochs} of  personalization (High Computation regime) starting from prompts pre-trained  by FedAvg(Adam), FedAvg(SGD) and FedSGD with random seed 1 on 
    \textbf{(Left)} HHF-SNI,   \textbf{(Center)} MHF-SNI, and  \textbf{(Right)} LHF-SNI.
    }
    \label{fig:seed1}
\end{figure}

\textbf{Variation across training runs.} In Figures \ref{fig:seed0} and \ref{fig:seed1} we plot versions of Figure \ref{fig:fl} with different random seeds for training. In each case the takeaway is the same as Observations 3a,b: FedAvg(Adam) outperforms FedAvg(SGD), and FedAvg(Adam) generally outperforms FedSGD, especially when trained on low-heterogeneity data and especially in terms of global scores. The one case in which FedSGD yields a better personalization-robustness tradeoff is on HHF-SNI (high heterogeneity) with seed 0 (Figure \ref{fig:seed0}) due to higher local scores for FedSGD.

\begin{figure}
    \centering
        \includegraphics[width=0.37\textwidth]{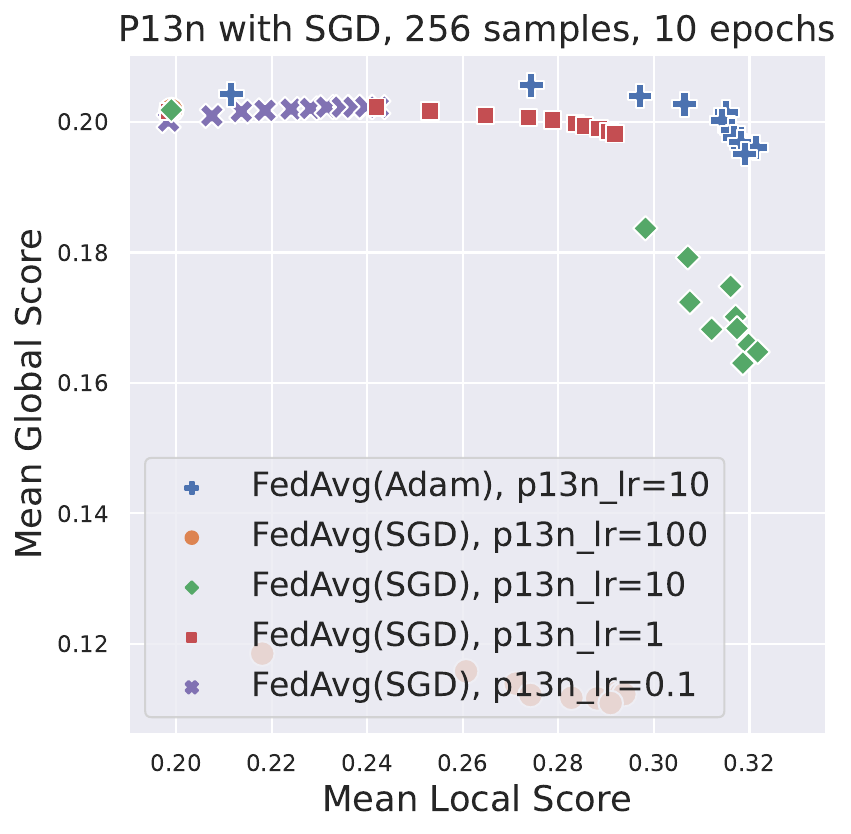}  
    \caption{\textbf{Personalization with SGD.}  Mean global and local scores across test clients evaluated every  epoch during 10 epochs of personalization with SGD, starting from prompts pre-trained by FedAvg(Adam) and FedAvg(SGD) on HHF-SNI. }
    \label{fig:sgd_p13n}
\end{figure}

\textbf{SGD as personalization optimizer.} One may suspect that the improvement of FedAvg(Adam) over FedAvg(SGD) in the previous results is due to FedAvg(Adam) using the same client optimizer as the personalization optimizer (Adam). However, Figure \ref{fig:sgd_p13n} we show that the relative performance of FedAvg(Adam) and FedAvg(SGD) does not change when SGD is used as the personalization optimizer rather than Adam.

\begin{figure}
    \centering
    \begin{subfigure}[b]{0.317\textwidth}
        \centering
        \includegraphics[width=\textwidth]{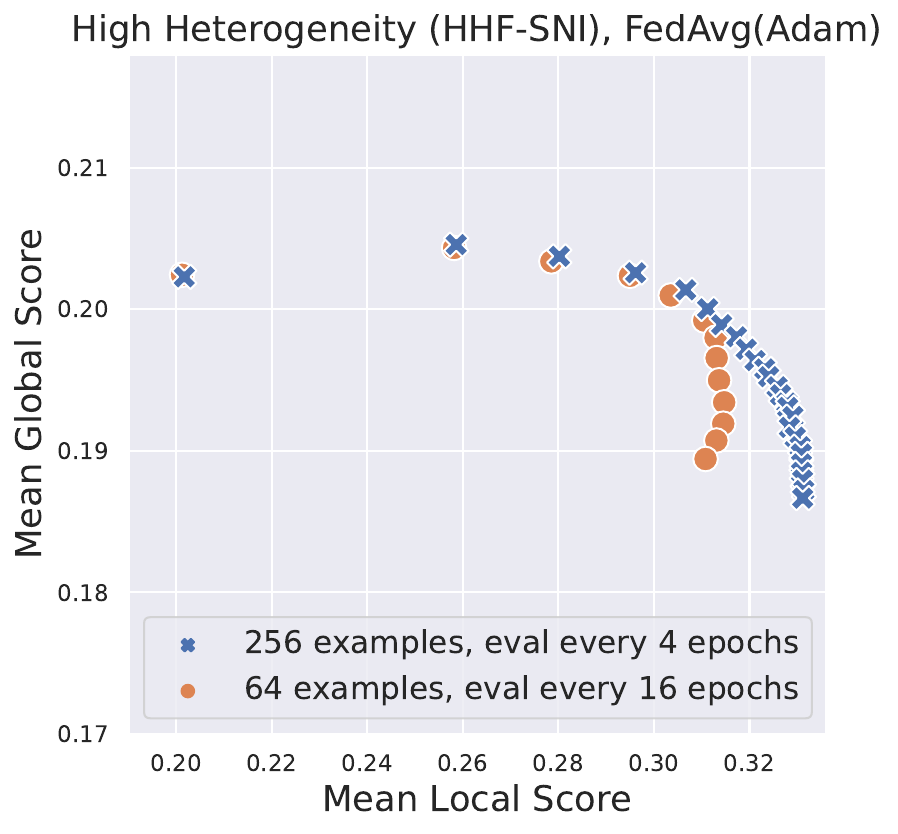}  
    \end{subfigure}
    \hfill
    \begin{subfigure}[b]{0.329\textwidth}
        \centering
        \includegraphics[width=\textwidth]{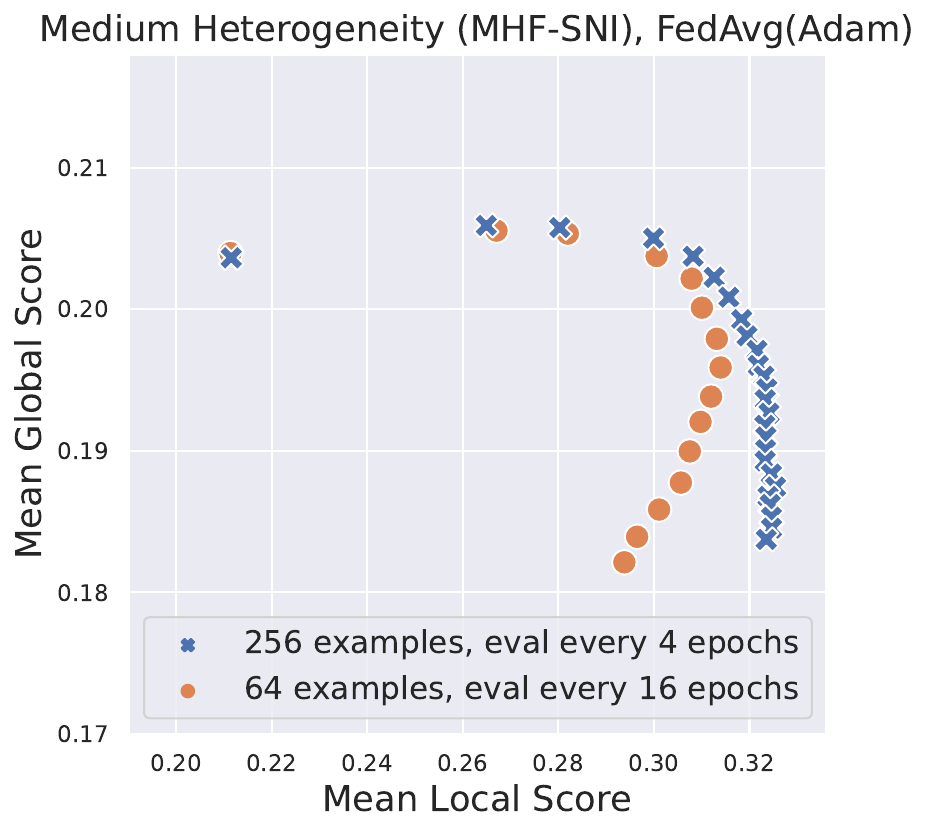}
    \end{subfigure}  
    \hfill
    \begin{subfigure}[b]{0.315\textwidth}
        \centering
        \includegraphics[width=\textwidth]{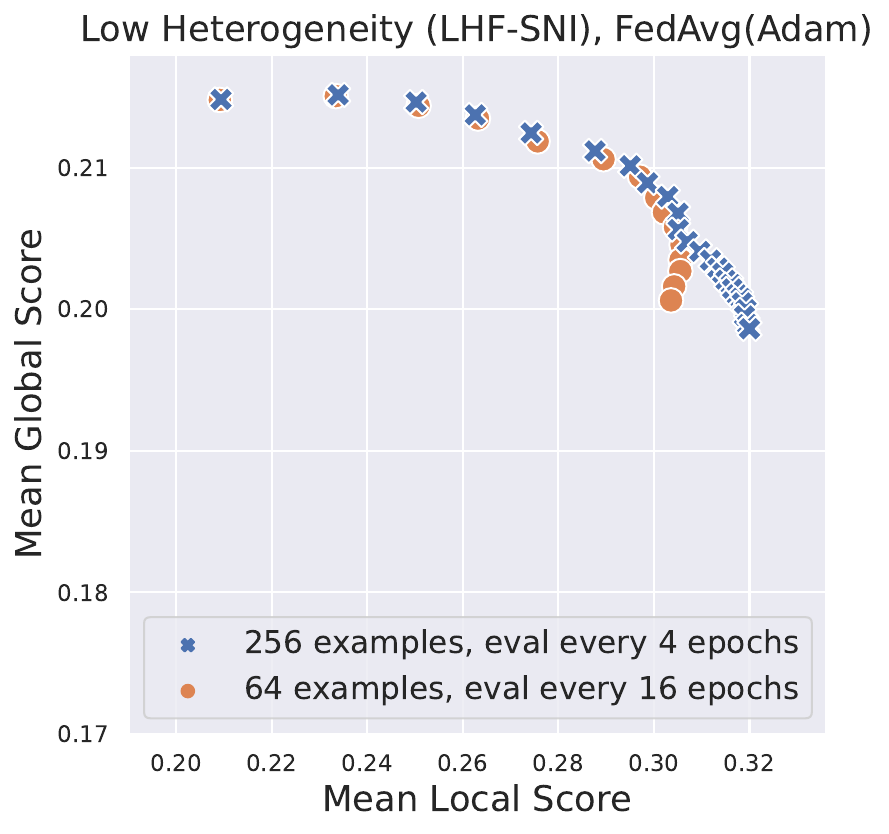}
    \end{subfigure}  
    \caption{\textbf{Impact of fewer personalization instances.} Global and local scores during personalization on either 256 or 64 instances (examples) starting from prompts pre-trained on \textbf{(left)} HHF-SNI,  \textbf{(center)} MHF-SNI, and \textbf{(right)} LHF-SNI. For 256 instances, 100 epochs are executed, and for 64 instances,  224 epochs are executed (High 
    Computation regime). 
    }
    \label{fig:num_samples}
\end{figure}

\begin{figure}
    \centering
    \begin{subfigure}[b]{0.32\textwidth}
        \centering
        \includegraphics[width=\textwidth]{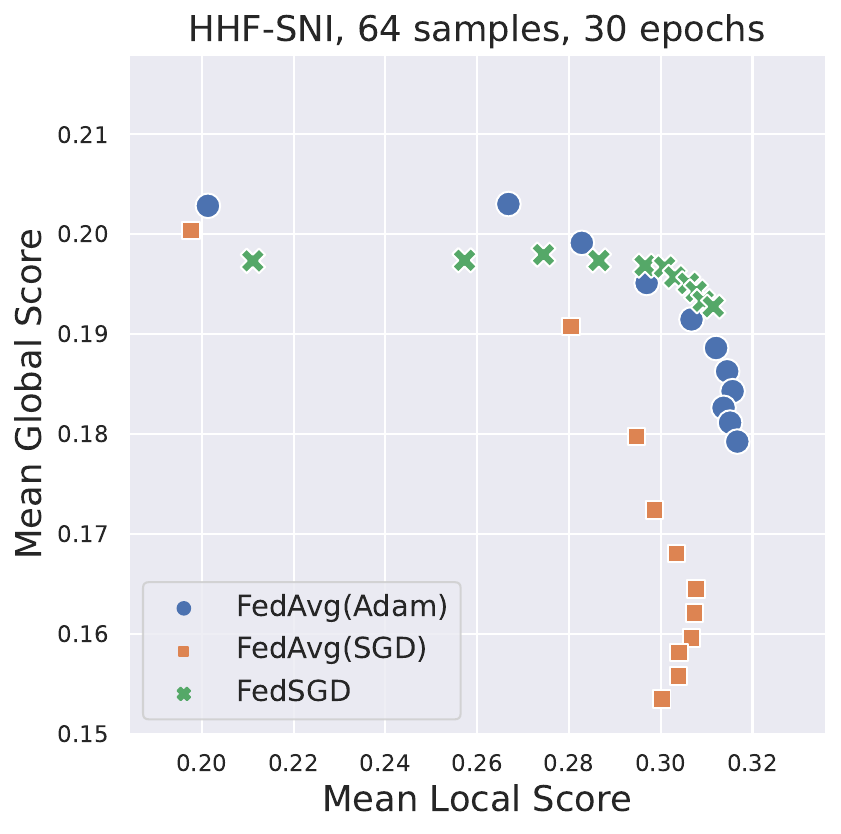}  
    \end{subfigure}
    \hfill
    \begin{subfigure}[b]{0.32\textwidth}
        \centering
        \includegraphics[width=\textwidth]{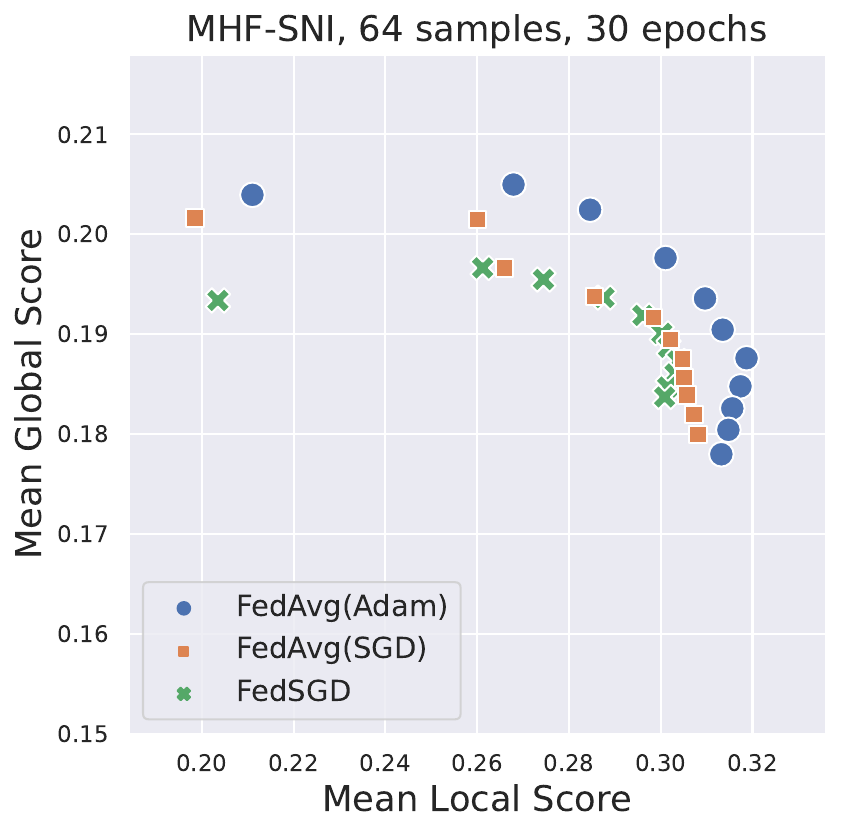}
    \end{subfigure}  
    \hfill
    \begin{subfigure}[b]{0.32\textwidth}
        \centering
        \includegraphics[width=\textwidth]{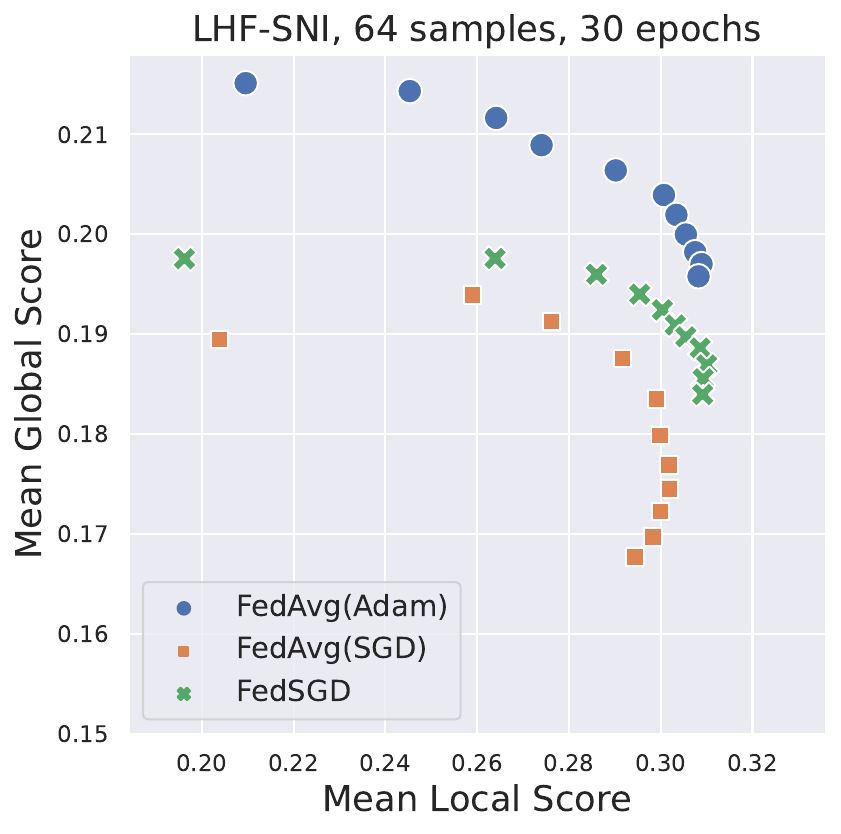}
    \end{subfigure}  
    \caption{\textbf{Low computation, 64 instances.} Mean global and local scores across test clients evaluated every 3 epochs during personalization with 30 total epochs of 64 instances (samples) per epoch, from prompts pre-trained on \textbf{(left)} HHF-SNI,  \textbf{(center)} MHF-SNI, and \textbf{(right)} LHF-SNI.}
    \label{fig:64-30}
\end{figure}

\textbf{Impact of fewer personalization samples.} 
In Figure \ref{fig:num_samples} we plot results from personalization with varying number of of examples per client, namely 64 and 256. With only 64 samples, late in training overfitting to the training set occurs to extent that even local scores decrease. Further, the best local score for 64 examples  is smaller than the best local score for 256 examples by about 0.01-0.02 for each heterogeneity level. However, fewer local samples reduces local scores more so than global scores, and early in training the personalization-robustness trade-off is roughly equivalent to  that with 256 examples.

In Figure \ref{fig:64-30}, we compare the personalization vs robustness trade-off for FedAvg(Adam), FedAvg(SGD), and FedSGD-trained prompts with few instances (64) in the Low Computation rage (30 epochs). Note that this is more updates than the previously studied Low Computation cases, which ran for 10 epochs, but the total amount of computation is actually less because we are here running epochs of 64 instances rather than 256 instances in the previous case. The relative ordering of performance among the three FL algorithms stays the same, with the exception of  FedSGD arguably slightly outperforming FedAVg(Adam) in the heterogeneity case.

\begin{figure}
    \centering
    \begin{subfigure}[b]{0.32\textwidth}
        \centering
        \includegraphics[width=\textwidth]{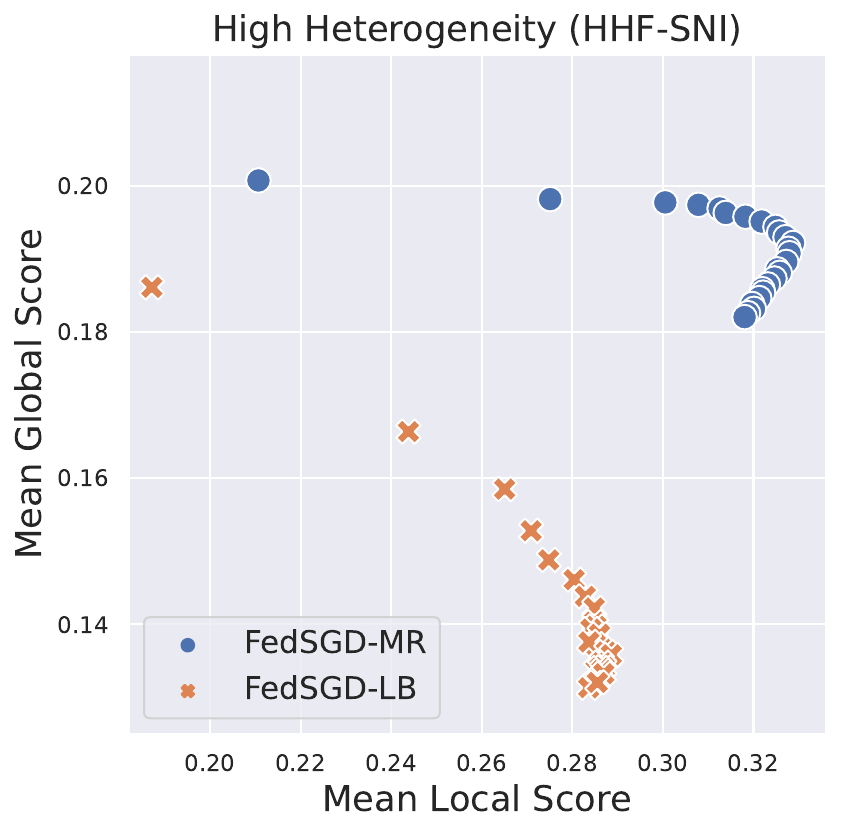}  
    \end{subfigure}
    \hfill
    \begin{subfigure}[b]{0.32\textwidth}
        \centering
        \includegraphics[width=\textwidth]{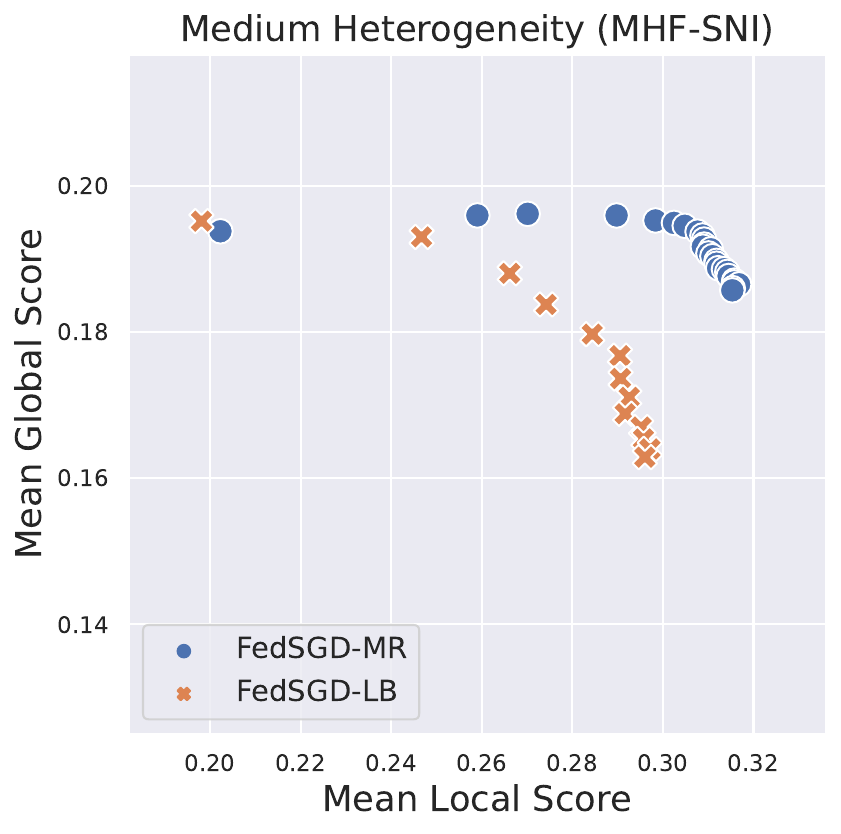}
    \end{subfigure}  
    \hfill
    \begin{subfigure}[b]{0.32\textwidth}
        \centering
        \includegraphics[width=\textwidth]{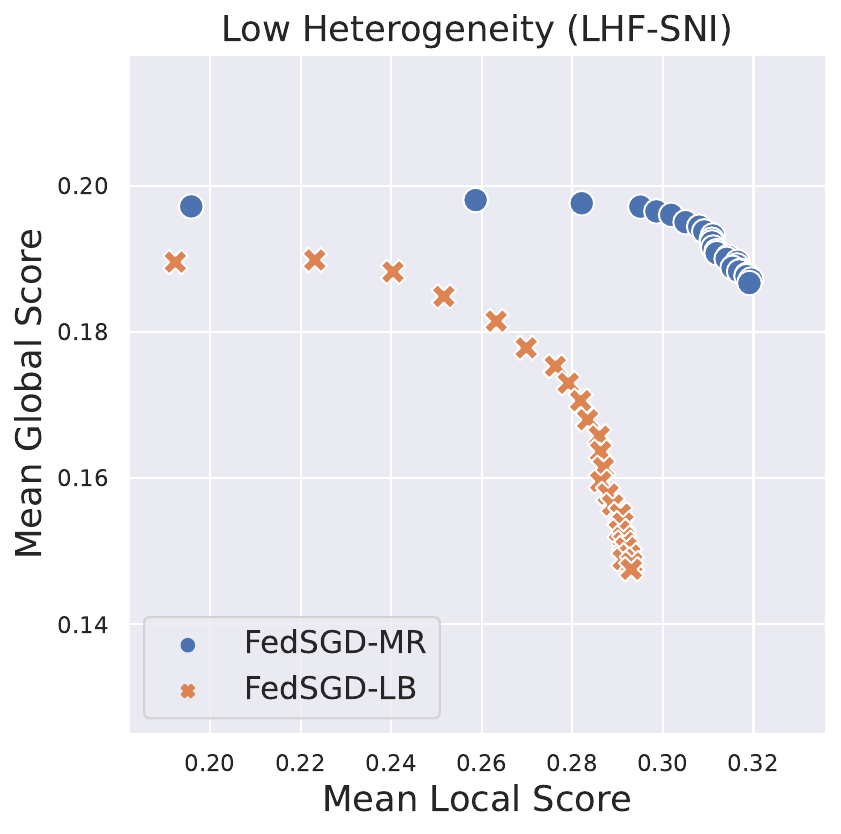}
    \end{subfigure}  
    \caption{\textbf{FedSGD with many rounds vs large batch size.}   Mean global and local scores across test clients during personalization starting from prompts pre-trained by FedSGD with many rounds (FedSGD-MR, referred to as FedSGD in all other experiments) and FedSGD with large batch size (FedSGD-LB) on \textbf{(left)} HHF-SNI,  \textbf{(center)} MHF-SNI, and \textbf{(right)} LHF-SNI. }
    \label{fig:fedsgd}
\end{figure}

\textbf{Variants of FedSGD.} In all previous experiments we have used the version of FedSGD that has the same client batch size (32) and number of active clients per round (32) as the FedAvg variants we experiment with, but executes 16x more communication rounds than the FedAvg variants (4800 rounds vs 1600 rounds) so that it sees the same total number of instances (since the FedAvg variants make 16 local updates per client per round, whereas  FedSGD makes effectively only 1). Now, we experiment with a different version of FedSGD that multiplies the client batch size by 16 rather than the number of communication rounds. In particular, this version, which we call \textbf{FedSGD-L}arge\textbf{B}atch, uses a client batch size of 512, and samples 32 clients per round for 300 rounds. Like the other FL algorithms, it uses Adam as its server optimizer. Figure \ref{fig:fedsgd} shows that the original version of FedSGD with \textbf{m}any \textbf{r}ounds (referred to here as FedSGD-MR) far outperforms FedSGD-LB, implying that it is advantageous to do more updates with noisier gradients.rather thank fewer updates with less noisy gradients.

\begin{figure}
    \centering
    \begin{subfigure}[b]{0.32\textwidth}
        \centering
        \includegraphics[width=\textwidth]{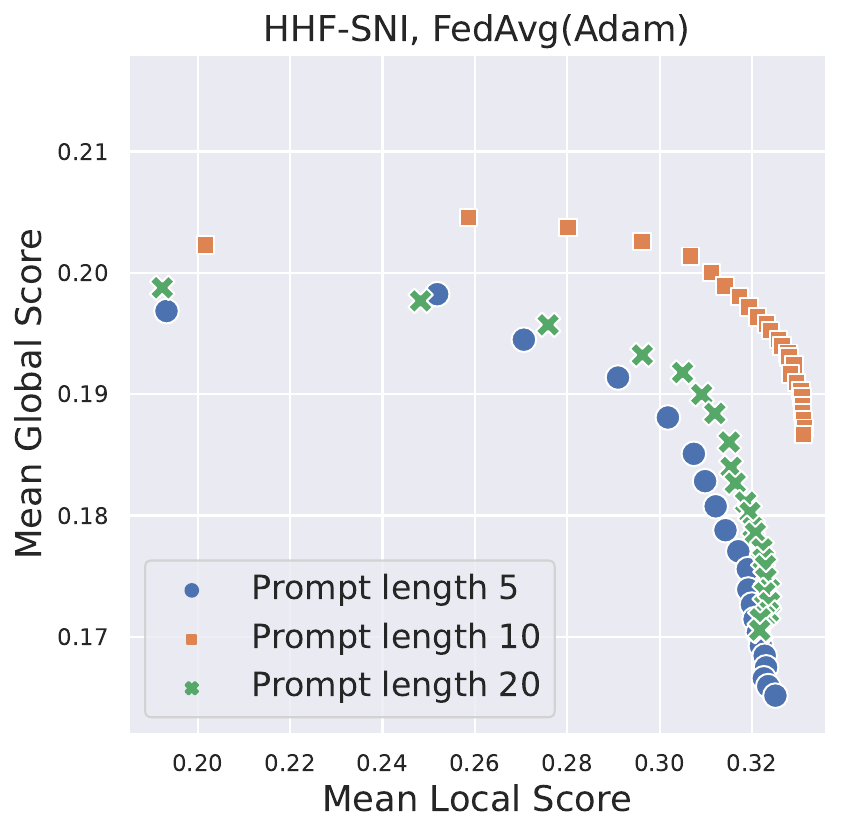}  
    \end{subfigure}
    \hfill
    \begin{subfigure}[b]{0.32\textwidth}
        \centering
        \includegraphics[width=\textwidth]{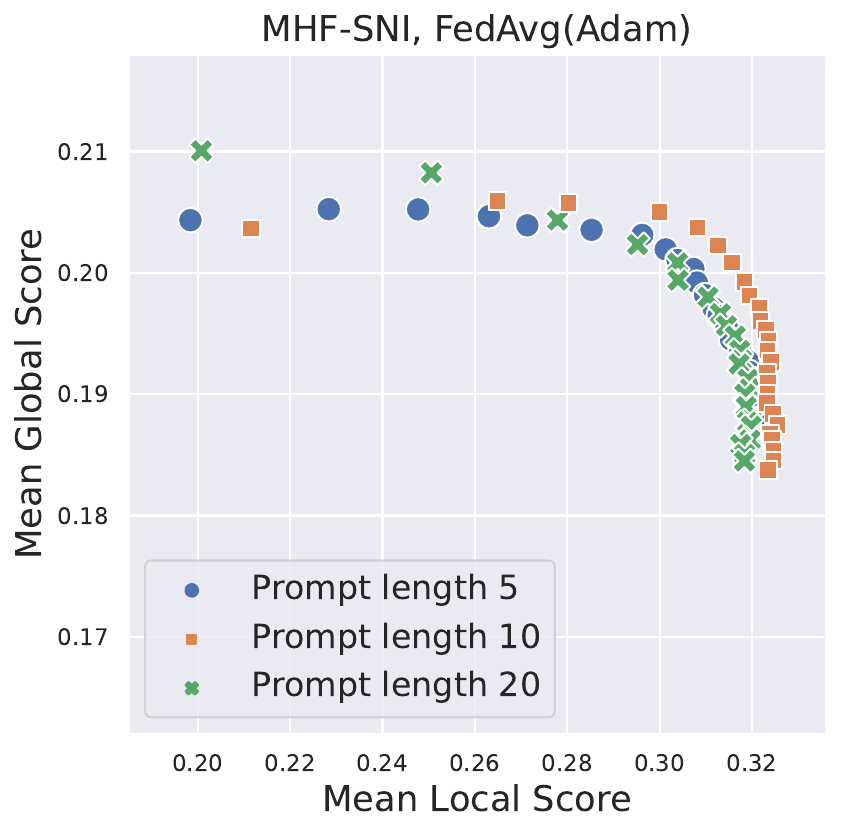}
    \end{subfigure}  
    \hfill
    \begin{subfigure}[b]{0.32\textwidth}
        \centering
        \includegraphics[width=\textwidth]{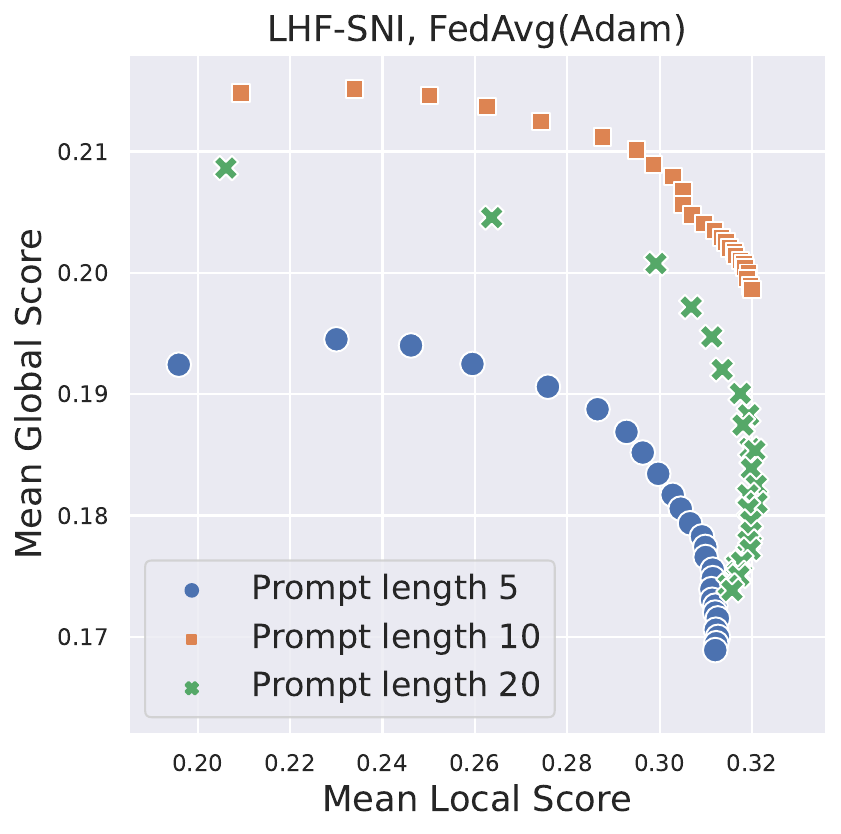}
    \end{subfigure}  
    \caption{\textbf{Role of prompt length -- FedAvg(Adam).}  Mean global and local scores evalutated every 4 epochs during 100 epochs of personalization on 256 instances starting from prompts of varying lengths pre-trained by FedAvg(Adam) on \textbf{(left)} HHF-SNI,  \textbf{(center)} MHF-SNI, and \textbf{(right)} LHF-SNI. }
    \label{fig:prompt_length_fedavg}
\end{figure}

\begin{figure}
    \centering
    \begin{subfigure}[b]{0.32\textwidth}
        \centering
        \includegraphics[width=\textwidth]{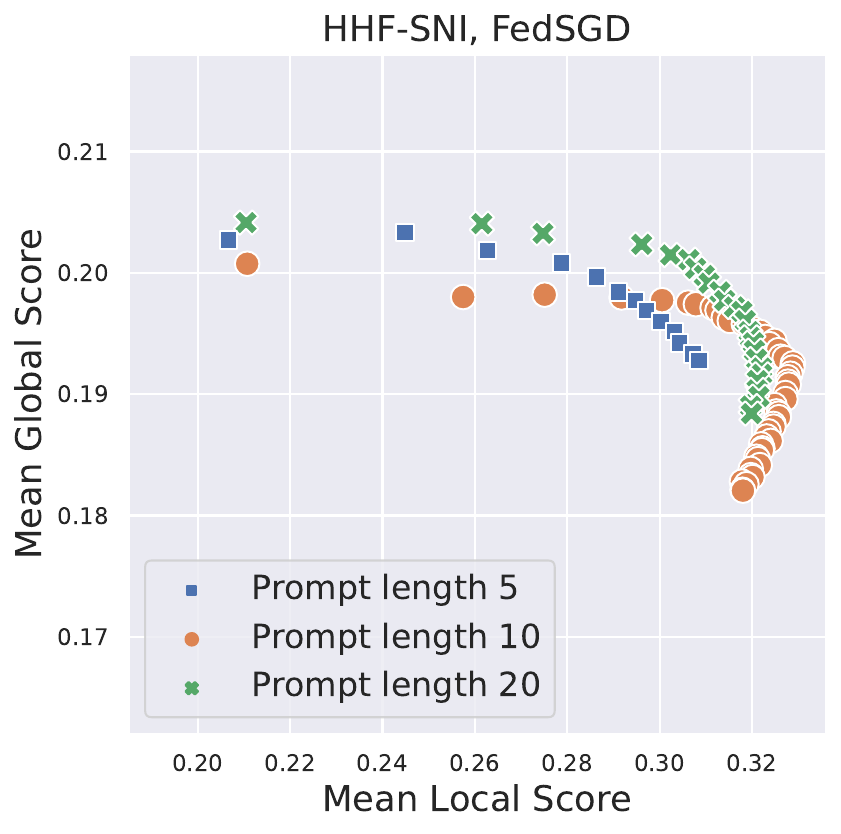}  
    \end{subfigure}
    \hfill
    \begin{subfigure}[b]{0.32\textwidth}
        \centering
        \includegraphics[width=\textwidth]{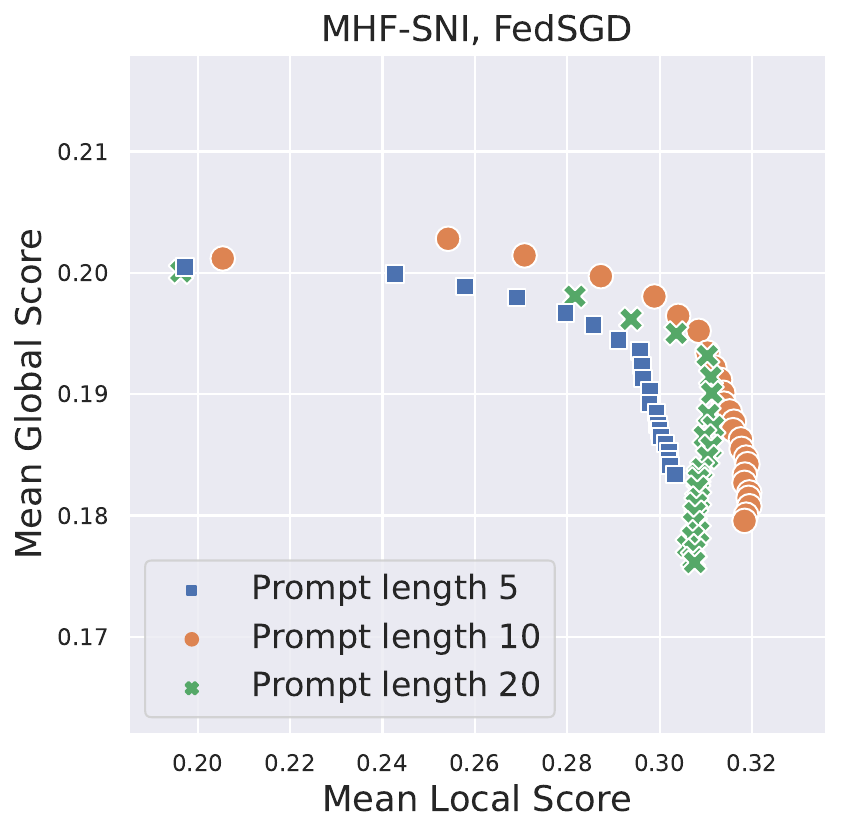}
    \end{subfigure}
    \hfill
    \begin{subfigure}[b]{0.32\textwidth}
        \centering
        \includegraphics[width=\textwidth]{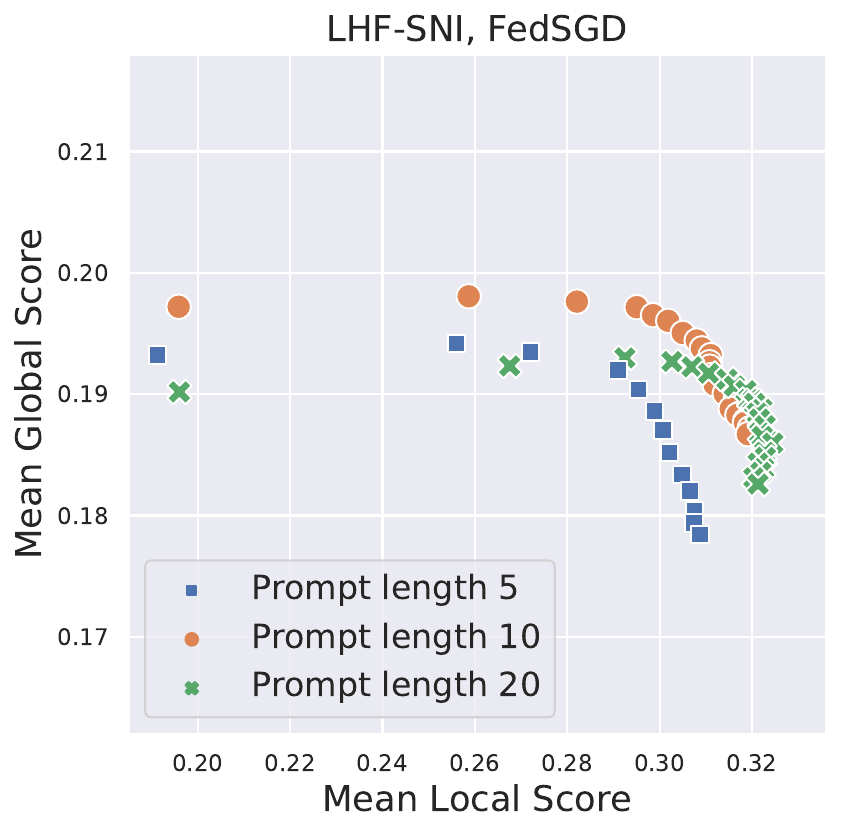}
    \end{subfigure}  
    \caption{\textbf{Role of prompt length -- FedSGD.}  Mean global and local scores evalutated every 4 epochs during 100 epochs of personalization on 256 instances starting from prompts of varying lengths pre-trained by FedSGD on \textbf{(left)} HHF-SNI,  \textbf{(center)} MHF-SNI, and \textbf{(right)} LHF-SNI. }
    \label{fig:prompt_length_fedsgd}
\end{figure}

\begin{figure}

    \centering
    \begin{subfigure}[b]{0.32\textwidth}
        \centering
        \includegraphics[width=\textwidth]{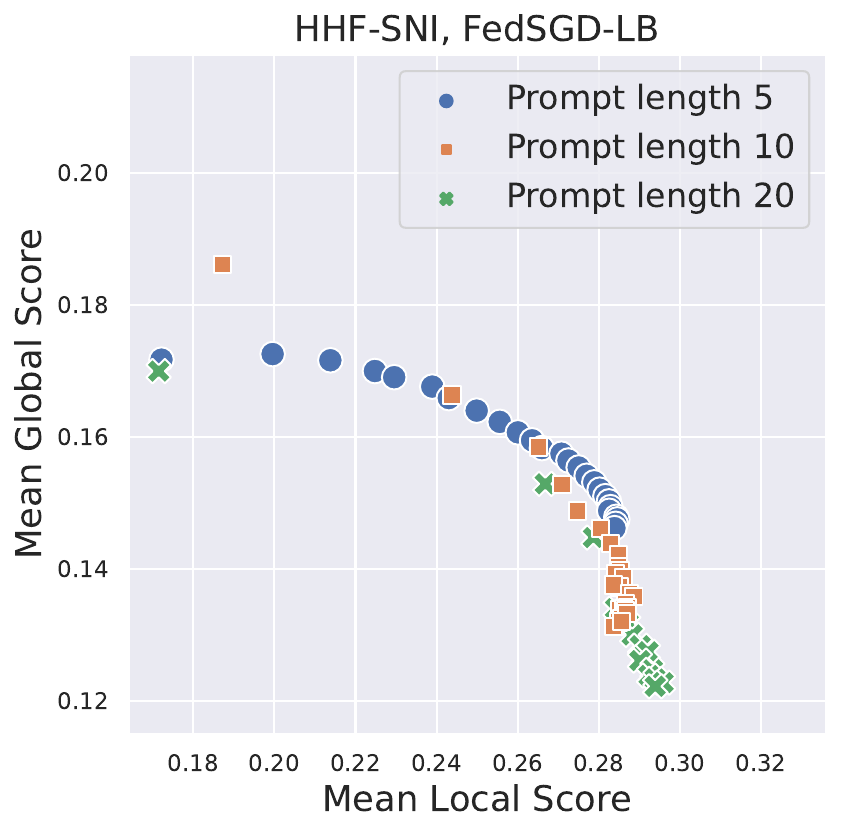}  
    \end{subfigure}
    \hfill
    \begin{subfigure}[b]{0.32\textwidth}
        \centering
        \includegraphics[width=\textwidth]{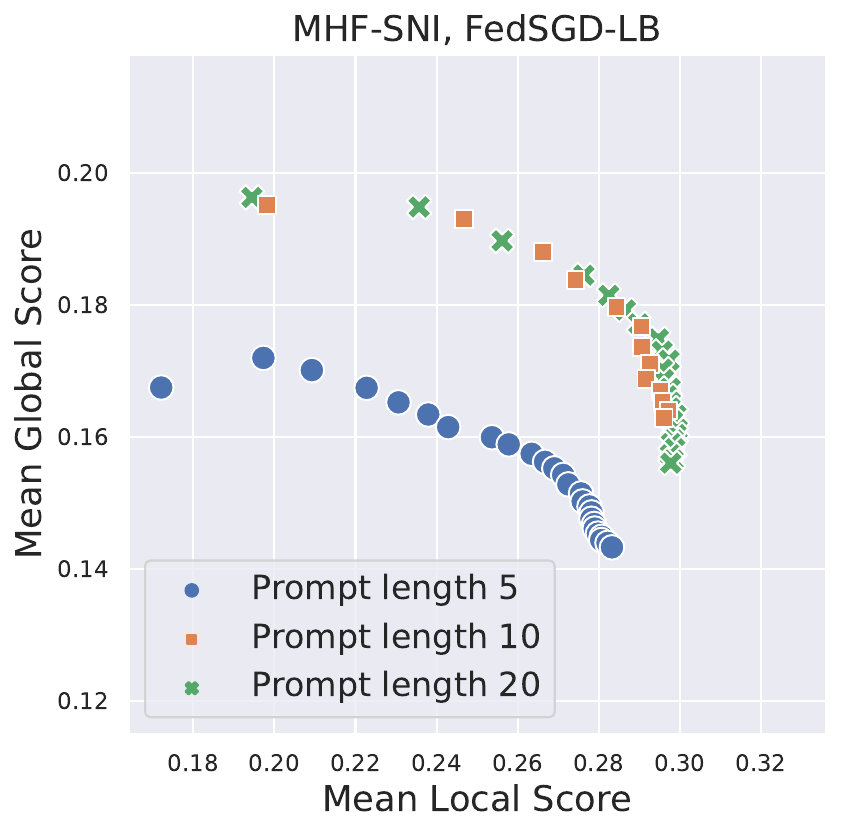}
    \end{subfigure}
    \hfill
    \begin{subfigure}[b]{0.32\textwidth}
        \centering
        \includegraphics[width=\textwidth]{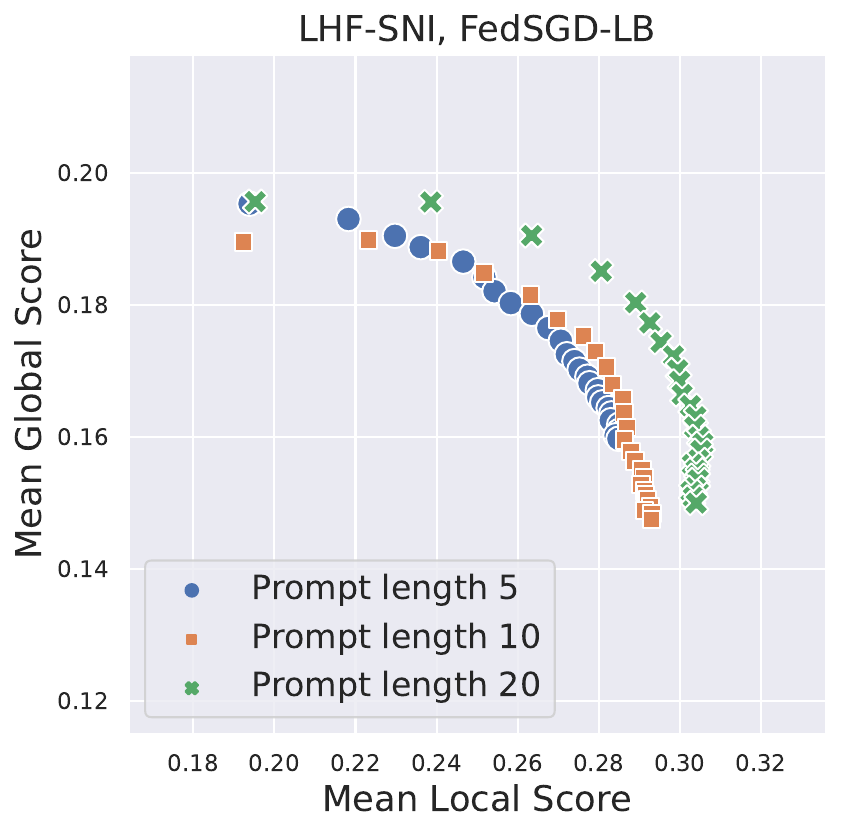}
    \end{subfigure}  
    \caption{\textbf{Role of prompt length -- FedSGD-LB.}  Mean global and local scores evalutated every 4 epochs during 100 epochs of personalization on 256 instances starting from prompts of varying lengths pre-trained by FedSGD-LB on \textbf{(left)} HHF-SNI,  \textbf{(center)} MHF-SNI, and \textbf{(right)} LHF-SNI. }
    \label{fig:prompt_length_fedsgd_lb}
\end{figure}

\textbf{Role of prompt length.} In Figures \ref{fig:prompt_length_fedavg}, \ref{fig:prompt_length_fedsgd} and \ref{fig:prompt_length_fedsgd_lb} we explore the effect of changing the prompt length for FedAvg(Adam), FedSGD and FedSGD-LB, respectively, in the High Computation personalization regime with 100 epochs of 256 samples. Prompt length 10 seems to be the sweet spot,  as prompt length 5 gives the worst personalization vs robustness trade-off in all cases besides FedSGD-LB on HHF-SNI, and prompt length 20 provides clear improvement over prompt length 10 only in one case (FedSGD-LB on LHF-SNI), and can sometimes do significantly worse (as in the FedAvg(Adam) cases). The  takeaway is similar to that in \cite{lester2021power}: increasing the number of tokens in soft prompts improves performance up to some number of tokens, but beyond this there is no benefit to further increasing the prompt length. 

\begin{figure}
\centering
        \includegraphics[width=0.37\textwidth]{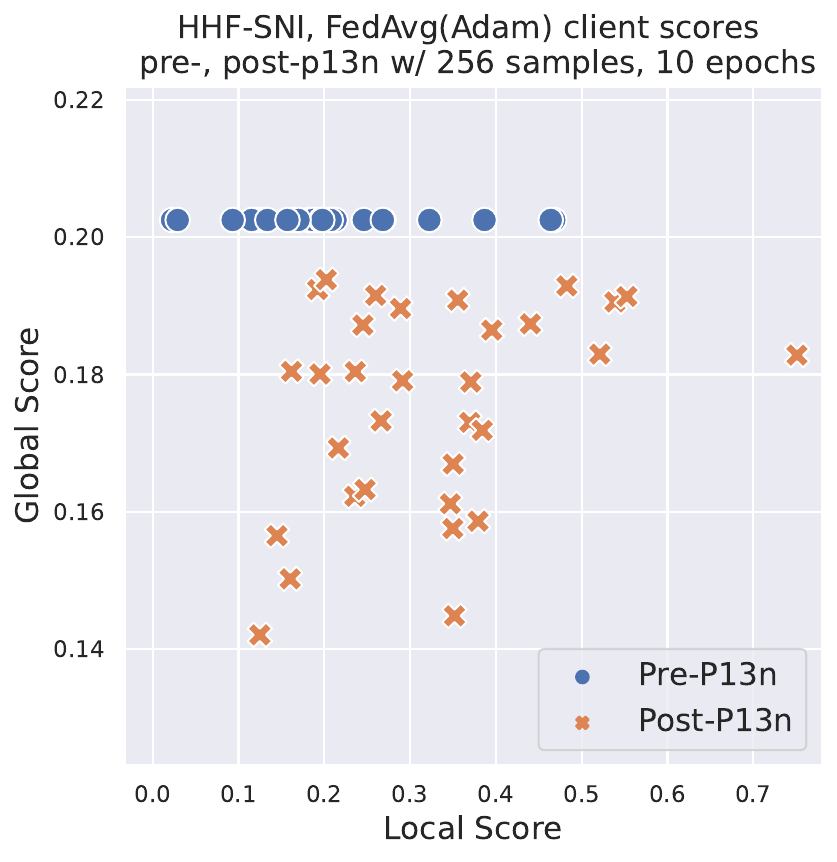} 
         
    \caption{Per-client  global and local scores before and after personalization (p13n) consisting of 10 epochs on 256 examples from prompts pre-trained by FedAvg(Adam) on HHF-SNI. }
    \label{fig:clients}
\end{figure}

\textbf{Variation in client performance.} Thus far all of our results have been mean scores across 32 test clients. Now, we investigate the variation in performance across clients. In  Figure \ref{fig:clients}, we plot each of the 32 test clients' scores pre- and post-personalization in the Low Computation regime with 10 epochs of personalization on 256 instances, starting from prompts trained by FedAvg(Adam) on  HHF-SNI. With the exception of one outlying client, the width of the range of local scores is roughly equivalent before and after personalization, while there is a large variance in global scores post-personalization.

\begin{figure}

    \centering
    \begin{subfigure}[b]{0.32\textwidth}
        \centering
        \includegraphics[width=\textwidth]{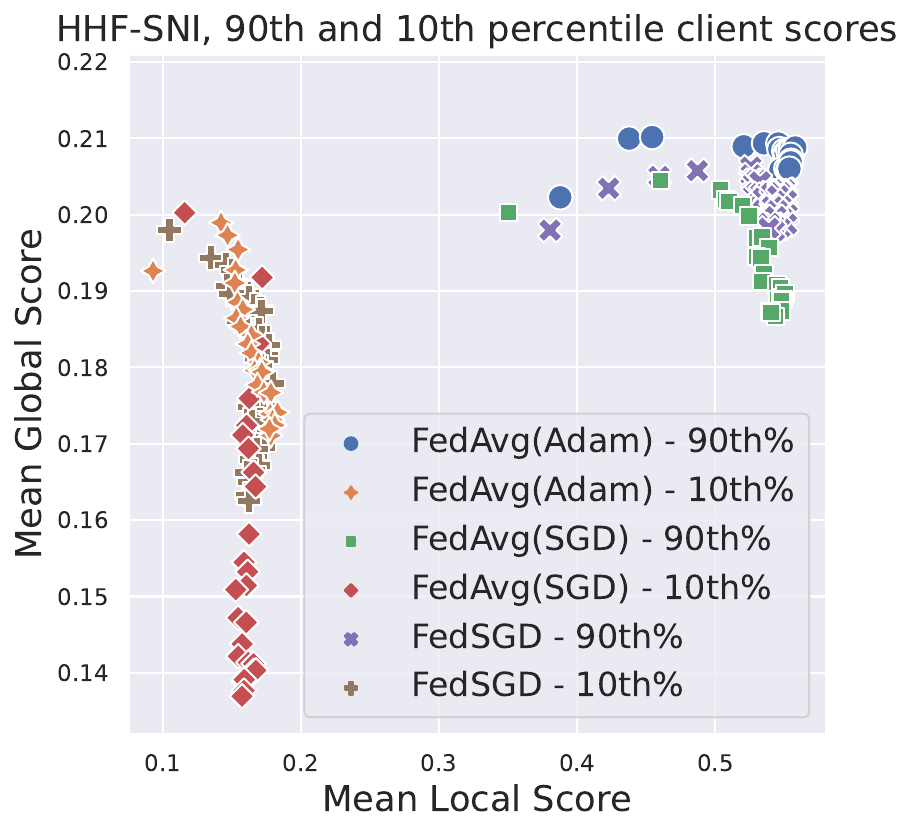}  
    \end{subfigure}
    \hfill
    \begin{subfigure}[b]{0.32\textwidth}
        \centering
        \includegraphics[width=\textwidth]{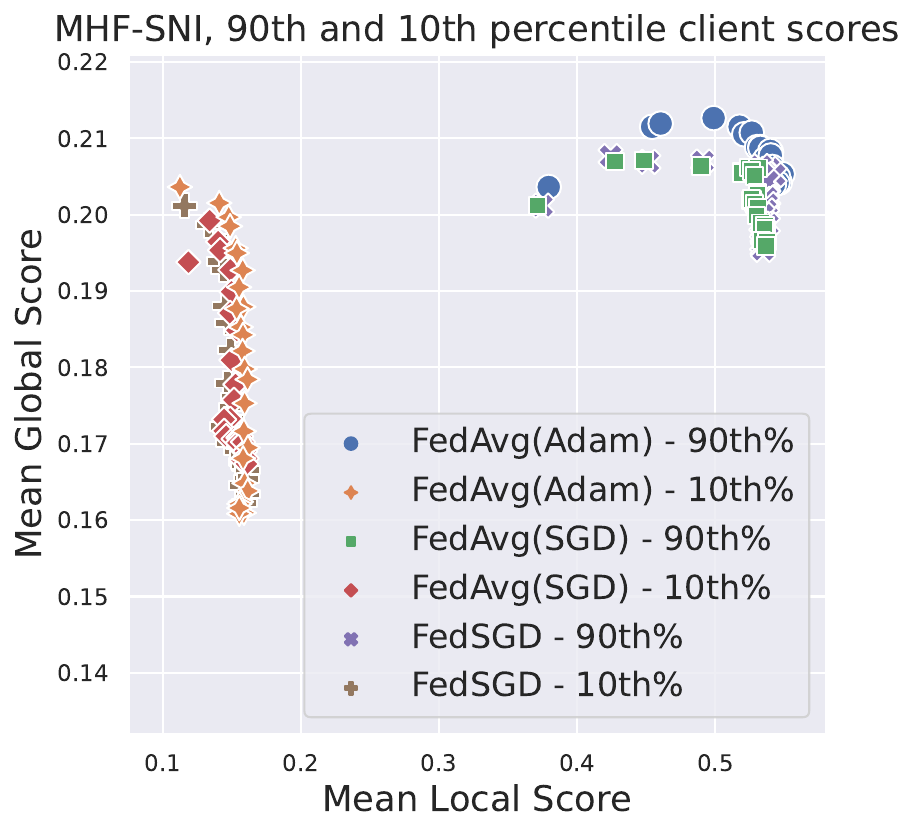}
    \end{subfigure}
    \hfill
    \begin{subfigure}[b]{0.32\textwidth}
        \centering
        \includegraphics[width=\textwidth]{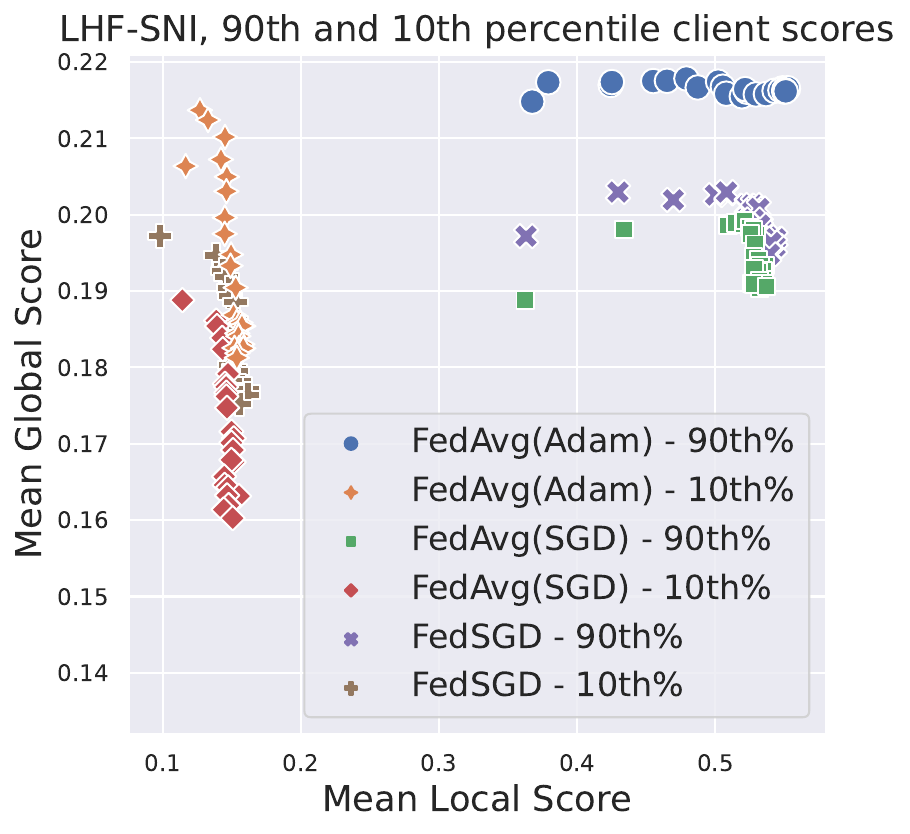}
    \end{subfigure}  
    \caption{Global and local  score 90th and 10th percentiles across test clients during personalization with 100 epochs of 256 instances from prompts pre-trained on \textbf{(left)} HHF-SNI,  \textbf{(center)} MHF-SNI, and \textbf{(right)} LHF-SNI. Scores are evaluated every 4 epochs.}
    \label{fig:percentiles}
\end{figure}

In Figure \ref{fig:percentiles}, we plot 90th and 10th percentile client global and local scores during personalization in the High Computation regime with 100 epochs of 256 instances from prompts trained by FedAvg(Adam), FedAvg(SGD), and FedSGD. That is, instead of each point representing (mean local score, mean global score) across clients during some personalization epoch, they instead represent (90th percentile local score, 90th percentile global score) across clients during some personalization epoch (and likewise for the 10th percentile). This yields a number of takeaways: 1) The worst local scores are roughly the same for all algorithms and during all personalization epochs, indicating that there are some very hard clients; 2) for all algorithms, the worst global scores drop significantly during personalization; 3) in contrast, the best global scores do not change much during personalization, and the best local scores increase significantly.

\subsection{Additional personalization heuristics}



\begin{figure}
    \centering
    \begin{subfigure}[b]{0.32\textwidth}
        \centering
        \includegraphics[width=\textwidth]{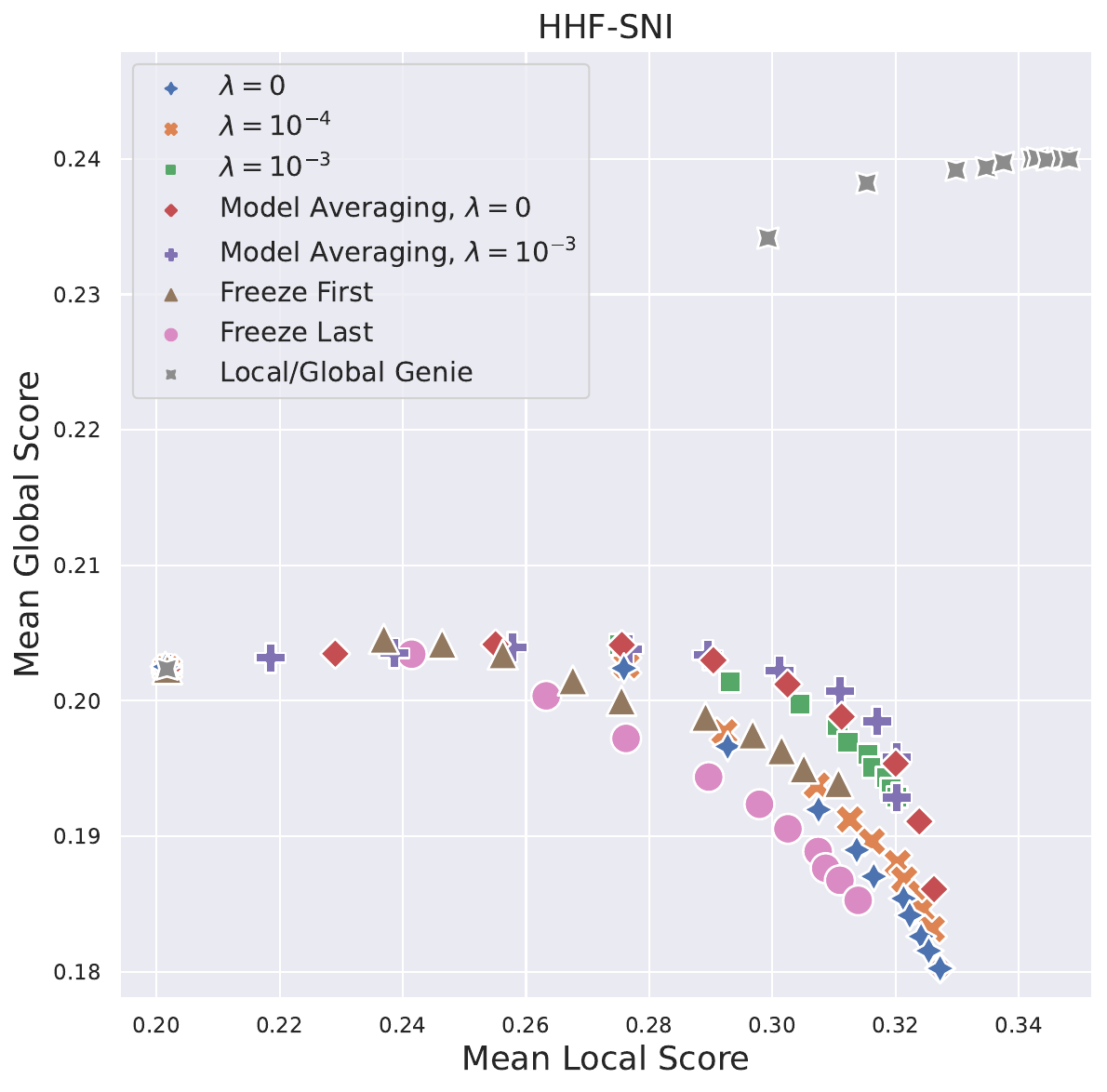}  
    \end{subfigure}
    \hfill
    \begin{subfigure}[b]{0.32\textwidth}
        \centering
        \includegraphics[width=\textwidth]{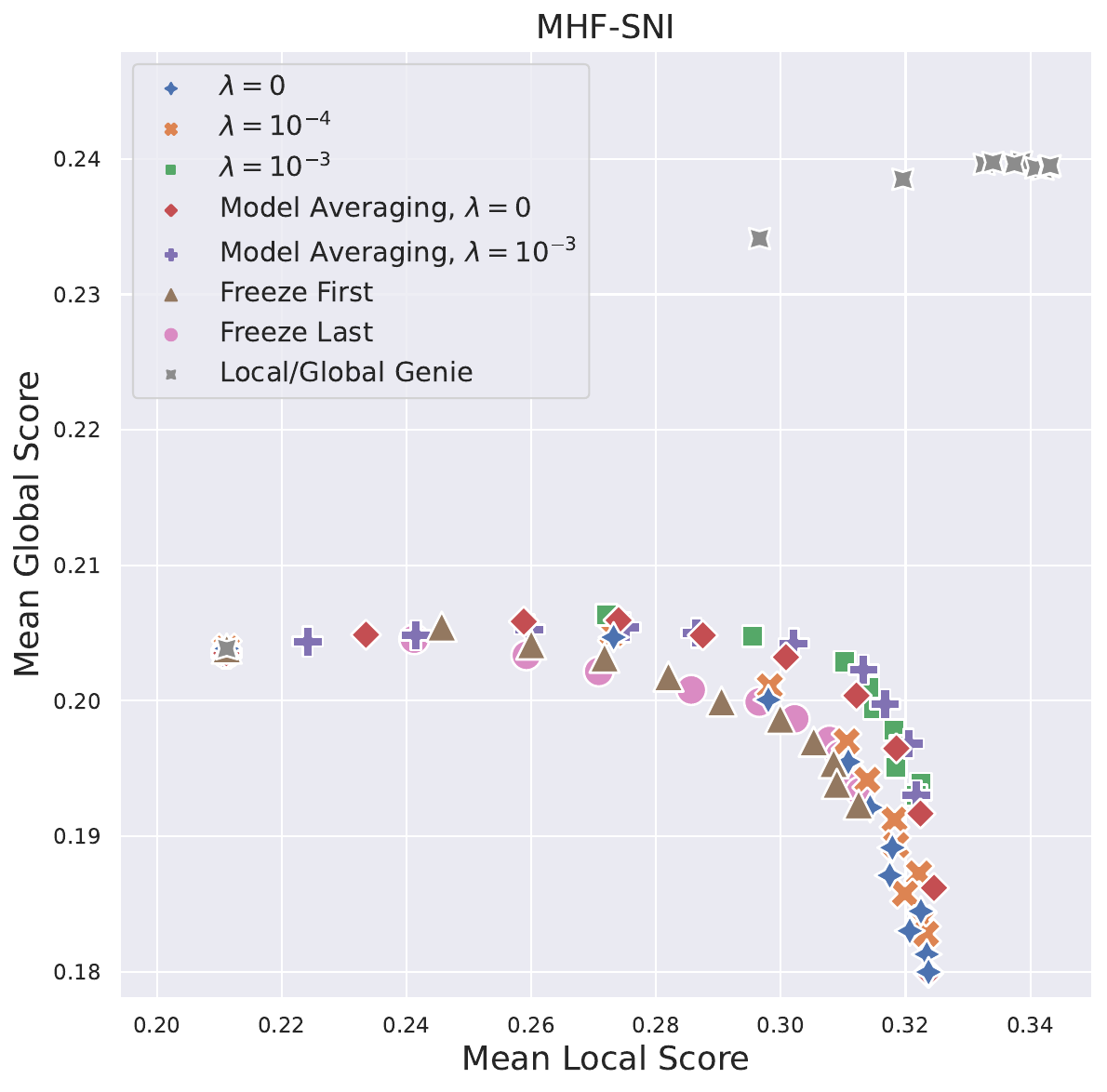}
    \end{subfigure}  
    \hfill
    \begin{subfigure}[b]{0.32\textwidth}
        \centering
        \includegraphics[width=\textwidth]{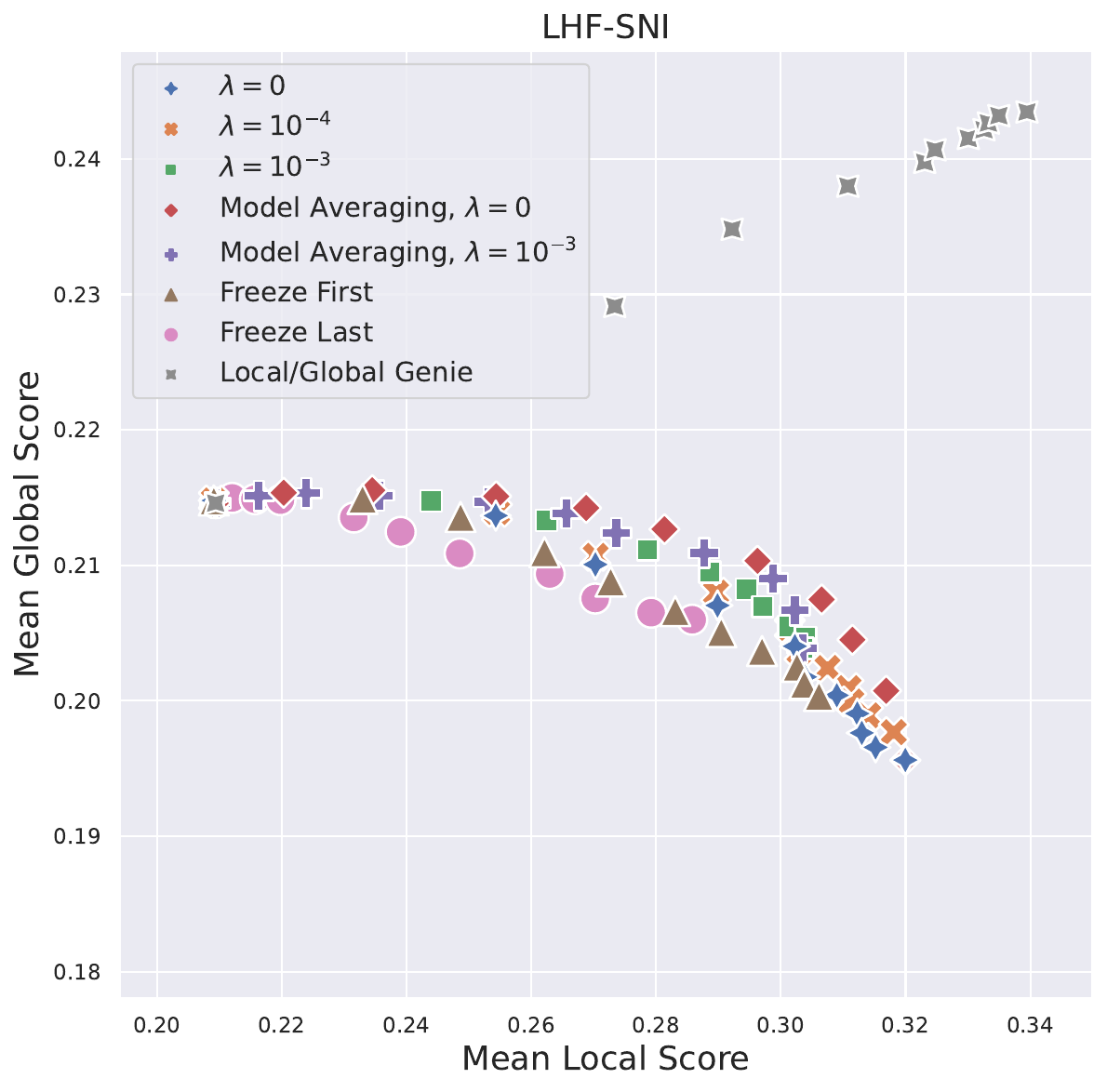}
    \end{subfigure}  
    \caption{
   \textbf{Additional personalization heuristics -- Low Computation.} Mean local and global scores during 10 epochs of personalization with various heuristics starting from prompts trained by FedAvg(Adam) on
    \textbf{(Left)} HHF-SNI,   \textbf{(Center)} MHF-SNI, and  \textbf{(Right)} LHF-SNI.
    }
    \label{fig:heuristics-full}
\end{figure}

Figure \ref{fig:heuristics-full} shows the same results as Figure \ref{fig:heuristics} plus results for three additional personalization approaches:
\begin{itemize}
    \item \textbf{Freeze First.} Recall that $P$ is a matrix of size prompt length (in tokens) by embedding dimension, where here the prompt length is 10. For ``Freeze First'', we freeze the first 8 rows (tokens) and only update the last two rows of $P_i$ (starting from $P_{\text{glob}}$) during personalization. 
    \item {\textbf{Freeze Last.}} Likewise, for ``Freeze Last'', we only update the first two rows of $P_i$. Neither ``Freeze First'' nor ``Freeze Last'' confer any improvement to the personalization-robustness trade-off.
    \item \textbf{Local/Global Genie.} These scores are the scores of a genie that knows the whether the personalized or global prompt will result in a prediction with larger score for a particular input and target, and uses the prompt with higher score for that input. It is equivalent to running inference twice for every input, once with the personalized prompt and once with the global prompt, and recording the max score among the two predictions, given the target. 
    This is not a realistic personalization method because in practice the target is unknown.  Nevertheless, we  find it to be a valuable measure of the combined capabilities of personalized and global prompts, i.e. the combined information between the personalized and global prompts. The very strong performance of this genie suggests that personalization-robustness trade-offs can be drastically improved by appropriately selecting whether to use the personalized prompt or global prompt for every input query (in fact, there would no longer be a trade-off -- both personalized and global scores would increase).
    To train the personalized prompt, we  we run vanilla personalization (i.e. $\lambda=0$ in Figure \ref{fig:heuristics-full}).
\end{itemize}

\end{document}